%% file: 01070.tex
\documentclass[final]{cvpr}

\usepackage{times}
\usepackage{epsfig}
\usepackage{graphicx}
\usepackage{amsmath}
\usepackage{amssymb}
\usepackage{mmstyle}
\usepackage{color}
\usepackage{booktabs}
\usepackage{multirow}
\newlength\savewidth\newcommand\shline{\noalign{\global\savewidth\arrayrulewidth
		\global\arrayrulewidth 1pt}\hline\noalign{\global\arrayrulewidth\savewidth}}

\usepackage[pagebackref=true,breaklinks=true,colorlinks,bookmarks=false]{hyperref}

\def\cvprPaperID{1070} 
\def\confYear{CVPR 2021}
\definecolor{orange}{RGB}{255,127,0}
\definecolor{blue1}{rgb}{0,0,1}
\definecolor{blue2}{rgb}{0,0.8,1}  

\begin{document}

\title{Scene-aware Generative Network for Human Motion Synthesis}

\author{Jingbo Wang$^{1}$
\quad \quad \quad Sijie Yan$^{1}$
\quad \quad \quad Bo Dai$^{2}$\thanks{Work done at The Chinese University of Hong Kong.}
\quad \quad \quad Dahua Lin$^{1,3}$ \\
$^{1}$CUHK - SenseTime Joint Lab, The Chinese University of Hong Kong \\
$^{2}$S-Lab, Nanyang Technological University\\
$^{3}$Centre of Perceptual and Interactive Intelligence\\
    {\tt\small \{wj020, ys016, dhlin\}@ie.cuhk.edu.hk},\hspace{5pt}
    {\tt\small \{bo.dai\}@ntu.edu.sg}   
}

\maketitle
\input{Sections/abstract}

\input{Sections/section_introduction_new}

\input{Sections/section_related_work}

\input{Sections/section_method}

\input{Sections/section_experiments}

\input{Sections/section_conclusion}
\input{Sections/appendix}

{\small
\bibliographystyle{ieee_fullname}
\bibliography{egbib}
}

\end{document}

%% file: Sections/abstract.tex
\begin{abstract}
We revisit human motion synthesis, a task useful in various real-world applications,
in this paper. 
Whereas a number of methods have been developed previously for this task, 
they are often limited in two aspects:
1) focus on the poses while leaving the location movement behind, and 
2) ignore the impact of the environment on the human motion.
In this paper, we propose a new framework, with the interaction between
the scene and the human motion taken into account.
Considering the uncertainty of human motion, we formulate this task
as a generative task, whose objective is to generate plausible human 
motion conditioned on both the scene and the human's initial position. 
This framework factorizes the distribution of human motions
into a distribution of movement trajectories conditioned on scenes and 
that of body pose dynamics conditioned on both scenes and trajectories.
We further derive a GAN-based learning approach, 
with discriminators to enforce the compatibility 
between the human motion and the contextual scene as well as 
the 3D-to-2D projection constraints.
We assess the effectiveness of the proposed method on two challenging 
datasets, which cover both synthetic and real-world environments.
\end{abstract}

%% file: Sections/section_introduction_new.tex
\section{Introduction}
\label{sec:intro}

The ability to synthesize human motions is beneficial to many real-world applications,
including virtual reality, filmmaking, and stochastic action forecasting.
Previous methods~\cite{barsoum2018hp,cai2018deep,fragkiadaki2015recurrent,chuan2020action2motion,  li2017auto,martinez2017human,yan2019convolutional,yang2018pose} for human motion synthesis often focus only on the movements of human bodies, while the scene context is neglected.
Basically, people move their bodies for interacting with the outside world and are restricted by the outside world.
It is hard to execute reasonable movements without observing the surrounding environment.
And thus, the problem is worth further exploring.

\begin{figure}[t]
\footnotesize
\centering
\renewcommand{\tabcolsep}{0pt} 
\begin{center}
\begin{tabular}{ccc}
\vspace{-3pt}
\includegraphics[width=0.33\linewidth]{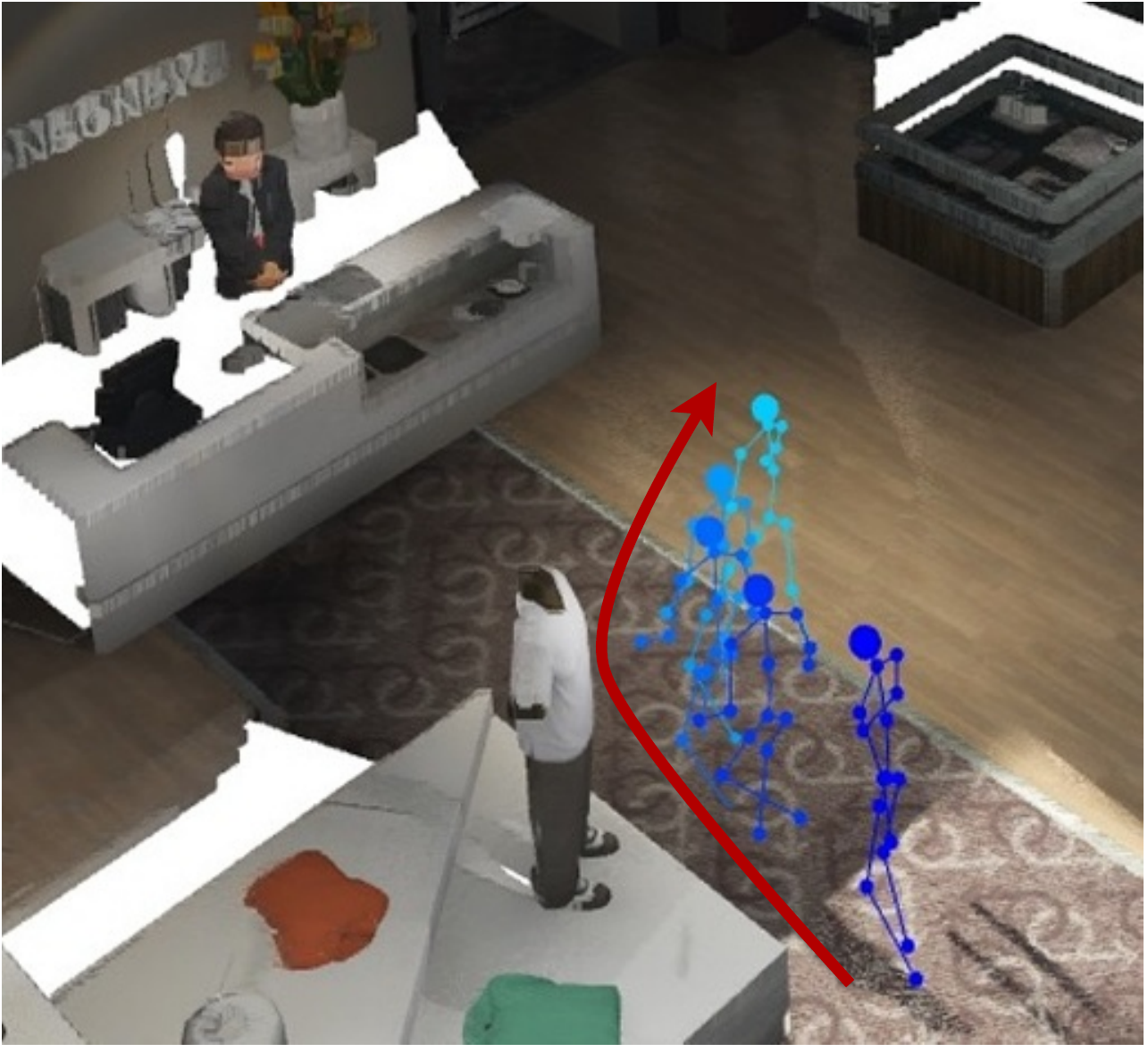} &
\includegraphics[width=0.33\linewidth]{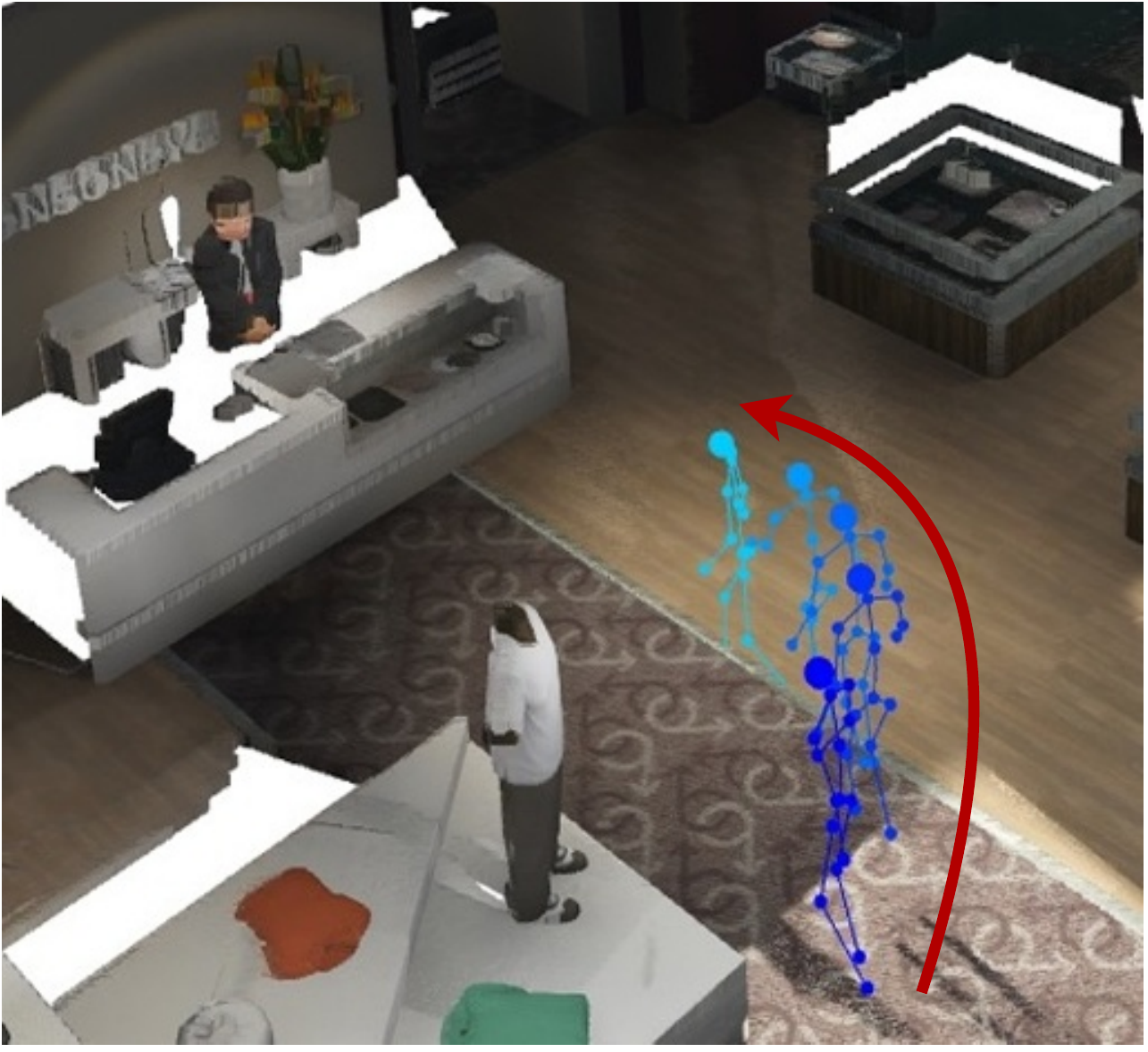} &
\includegraphics[width=0.33\linewidth]{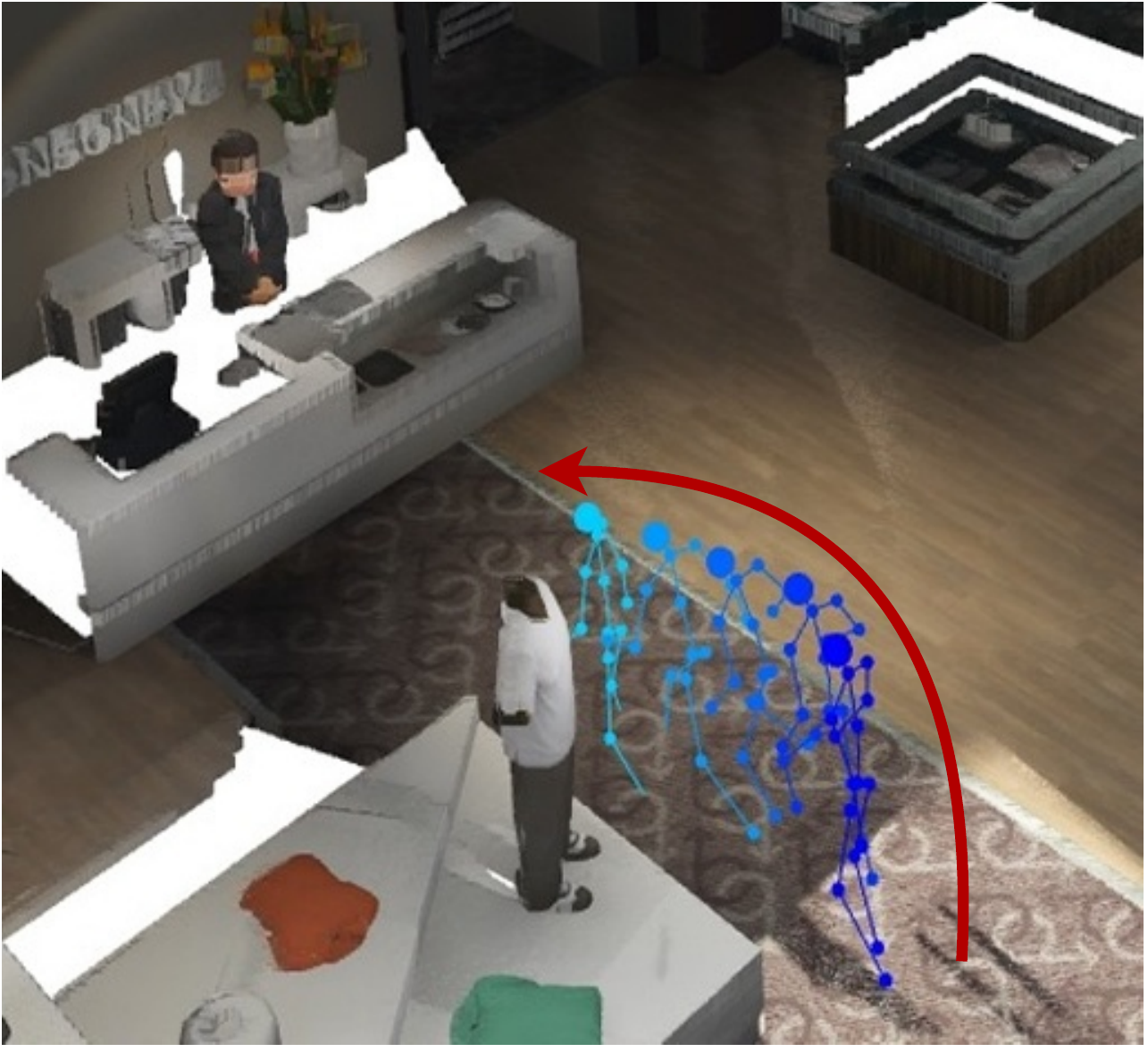} \\
\vspace{-3pt}
\includegraphics[width=0.33\linewidth]{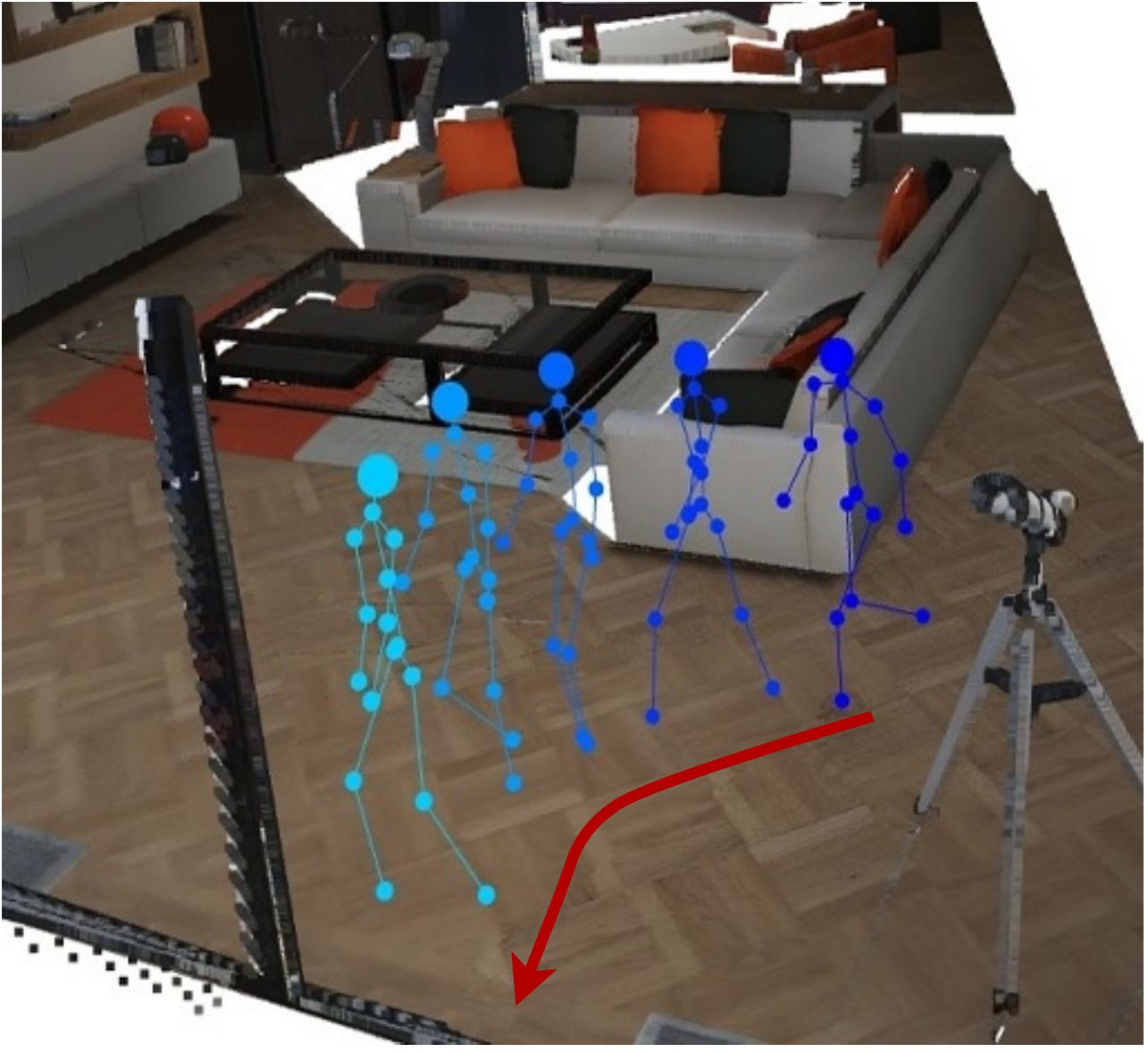} &
\includegraphics[width=0.33\linewidth]{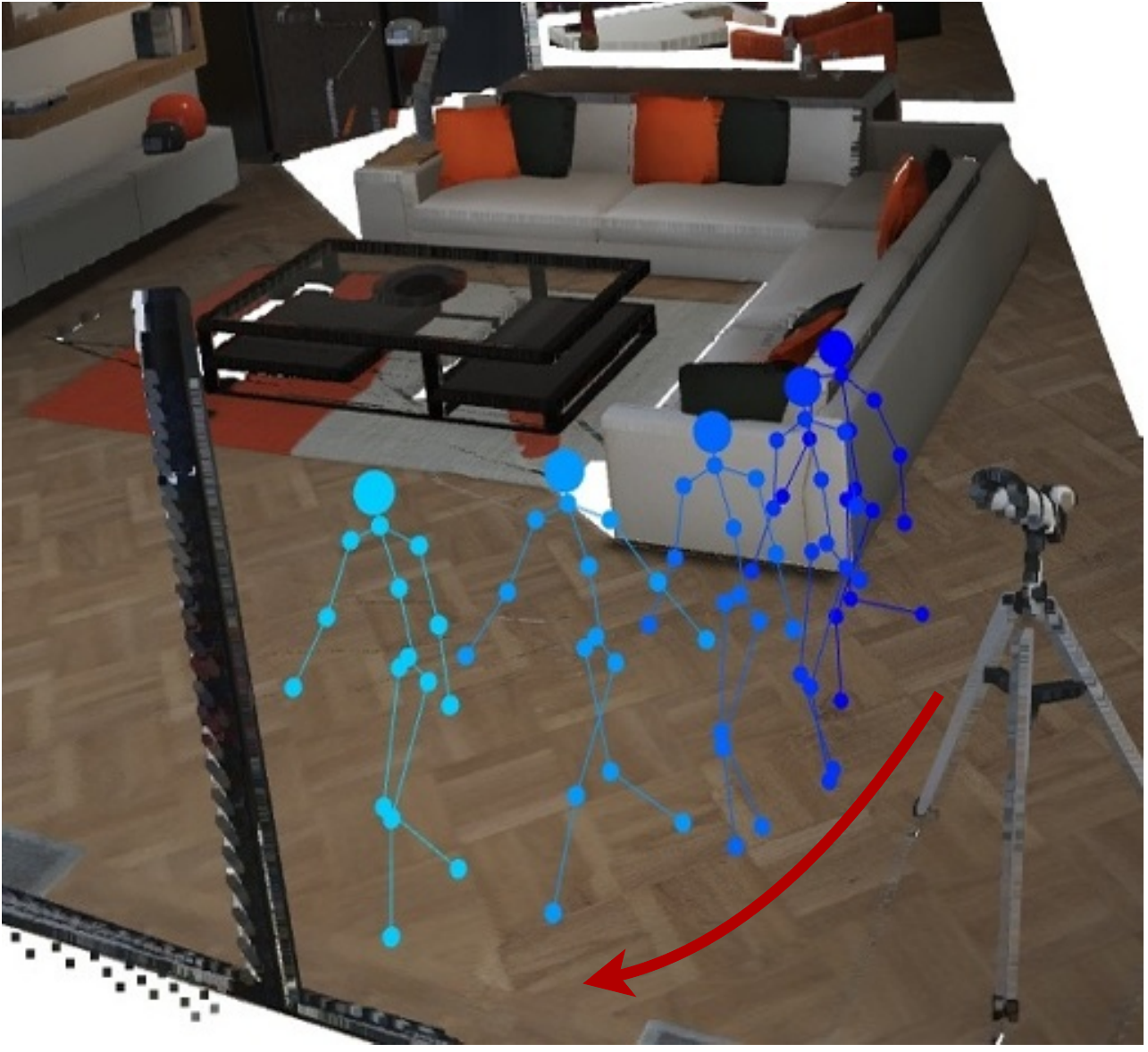} &
\includegraphics[width=0.33\linewidth]{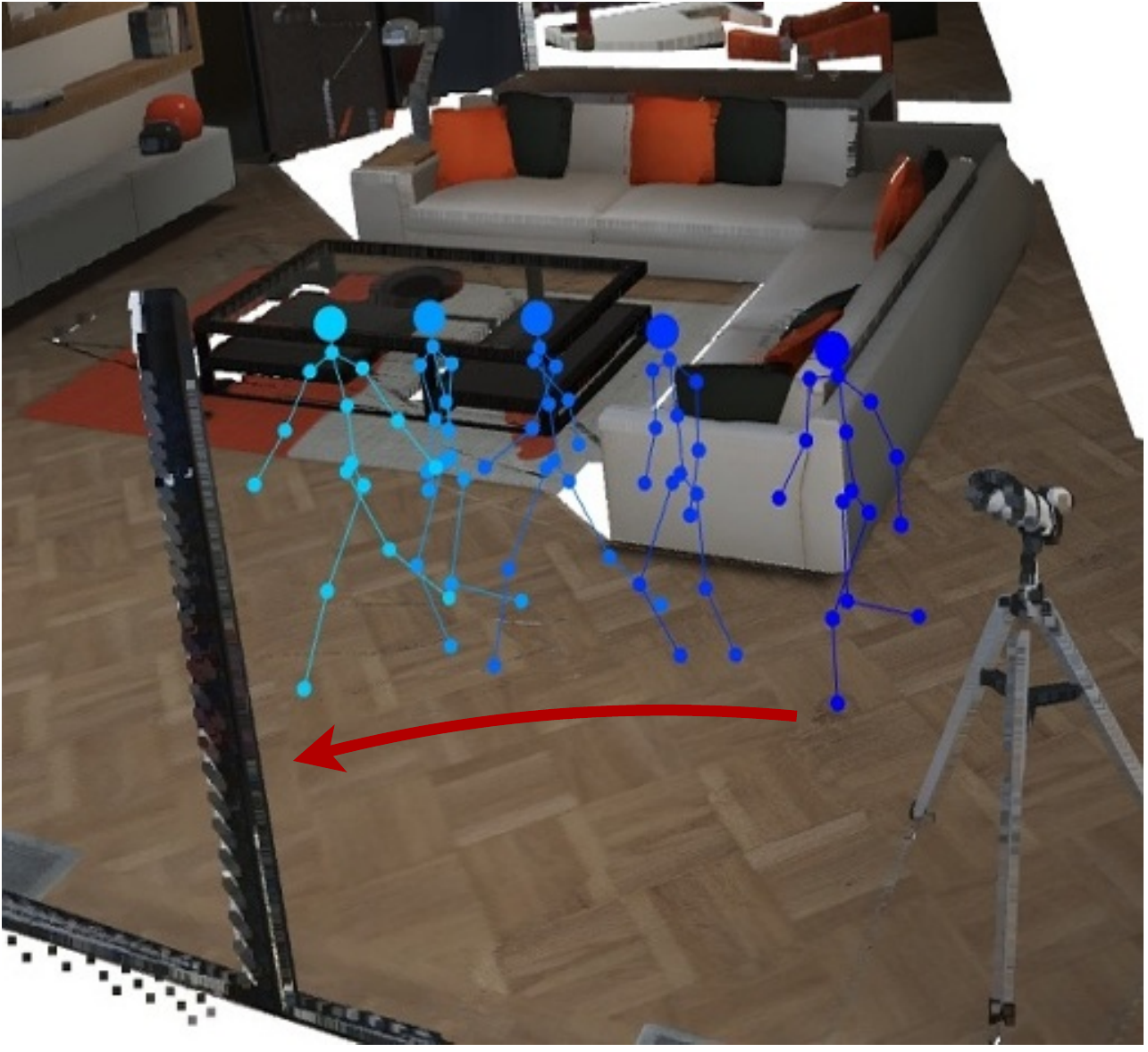} \\
\textbf{(a)} &
\textbf{(b)} &
\textbf{(c)}

\end{tabular}
\end{center}
\caption{\textbf{Visualization of human motions in different scenes}.
For the same \textcolor{blue1}{starting point}, the human not only can go to different \textcolor{blue2}{goals}, as in (a) and (c), but also the same \textcolor{blue2}{goals} under different trajectories and body movements, as in (a) and (b). All these human motions are sample from our generated results.}\vspace{-0.3cm}
\label{fig:teaser}
\end{figure}

Inspired by the importance of scene context, in this paper, we aim at synthesizing human motions under scene influence.
Actually, human motions in the scene consists of two components, namely body movements and the trajectory of human in the surrounding scene.
This trajectory controls the human movement in the scene, and the body movements always represent the action of humans, such as walking or sitting.
Thus, there are two major challenges to handle when involving scene context.
The first challenge is how to effectively reflect the semantic guidance provided by the scene context, \eg~do sitting action on a chair.
The second challenge is how to model the complicated physical relationship between scenes and action sequences. Specifically, we need to know the geometric configuration of the scene context to avoid the collision, \eg~where the floor is.

To solve these problems, there is an early attempt~\cite{cao2020long} introducing the scene context into motion forecasting, which supposes human actions are deterministic predictions when the history and the destination are given.
This method treats the distribution of human motion as the distribution of endpoints in the scene.
While in the motion synthesis task, we argue that such this treatment may lead to gaps between the learned distribution of human motion and the one in the real world since there could be infinite ways for a person to move from one place to another. They are all valid human motions, as shown in Figure~\ref{fig:teaser}.

Therefore, we propose our scene-aware fully generative framework to close this gap in motion synthesis.
This framework can learn the distribution of human motion in given scenes directly, rather than predicting human motions deterministically.
Following~\cite{cao2020long}, we represent scene context using an RGB image,
which is relatively easy to acquire in real scenarios.
Specifically, we divide the joint distribution as the trajectory prior in the scene and the conditional distribution of body movements given a trajectory.
Inspired by the success of convolutional sequence generation networks (CSGN)~\cite{yan2019convolutional} in skeleton-based action synthesis,
we introduce the scene context into CSGN to respectively model the trajectories and fine-grained body movements.
The distribution of trajectories is first learned by the trajectory generator under the condition of given scenes·
Intuitively, with the guidance of the scene context and the trajectory, it is easier for the pose generator to model the distribution of semantic compatible body movements than direct synthesis (\eg~human always do sitting action with the static trajectory and context information of chair).
In this way, our method is fully generative and is capable of capturing the diversity of human motions at various levels.

To fulfill the physical compatibility, it is crucial to introduce the geometry structure of the scene as prior knowledge into our synthesis framework.
Therefore, we supervise the encoder to extract geometry context from the scene by the depth map.
Under this supervision, the encoder can provide geometry-aware features of the scene, and we do not need to provide depth information of the scene during inference, which can be used more easily.
Moreover, we also propose a projection discriminator and a context discriminator as geometry discriminators to further encourage the compatibility between synthesized human motions and the surrounding scene context.
We deploy the projection discriminator on the 2D human motion in image coordinate space projected by the 3D human motion, because the abnormal human motion can be clearly exhibited by scale changes of humans in 2D space with the scene, as shown in Figure~\ref{fig:motion_projection}.
Therefore, this discriminator encourages the generator to synthesize trajectories following the global structure of the scene, such as the floor of the scene.
To prevent the collision between the synthesized human motions to the objects in the scene, we deploy this context discriminator to the relative depth sequence of the human motion to the local environment at each time step.
This discriminator encourages humans to move to the correct places surrounding these objects, which are shown in Figure~\ref{fig:context}.

We choose two challenging datasets to evaluate the effectiveness of
our proposed geometry-aware fully generative framework,
covering both a synthesized environment (\emph{GTA-IM}~\cite{cao2020long}) and a real environment (\emph{PROX}~\cite{PROX:2019}).
On both datasets,
the proposed framework is capable of synthesizing promising human motions,
in terms of the fidelity of each independent sequence,
the diversity of multiple sampled sequences,
as well as the consistency between synthesized sequences and their corresponding scenes.
To better quantitatively assess different methods,
we also propose a series of new metrics for human motion synthesis with scene context,
including Motion FID, which is inspired by FID for image synthesis,
and Non-collision Rate, which borrows insights from 3D computer games and examines the potential collisions between human motions and the scenes.

To summarize our contributions: 1) We reformulate the task of human motion synthesis with scene context as a conditional generation problem to avoid the limitations of deterministic prediction in previous works. We further propose a series of quantitative metrics to enhance the evaluation protocol of this task. 2) We develop a novel geometry-aware fully generative framework for this task, which explicitly takes the scene geometry into consideration and captures the diversity of human motions in a scene from multiple levels. 3)We propose two geometry-aware discriminators to encourage the compatibility between synthesized human motions and their corresponding scenes.

%% file: Sections/section_related_work.tex
\section{Related Works}

\begin{figure*}[t]
\centering
\includegraphics[width=0.9\linewidth]{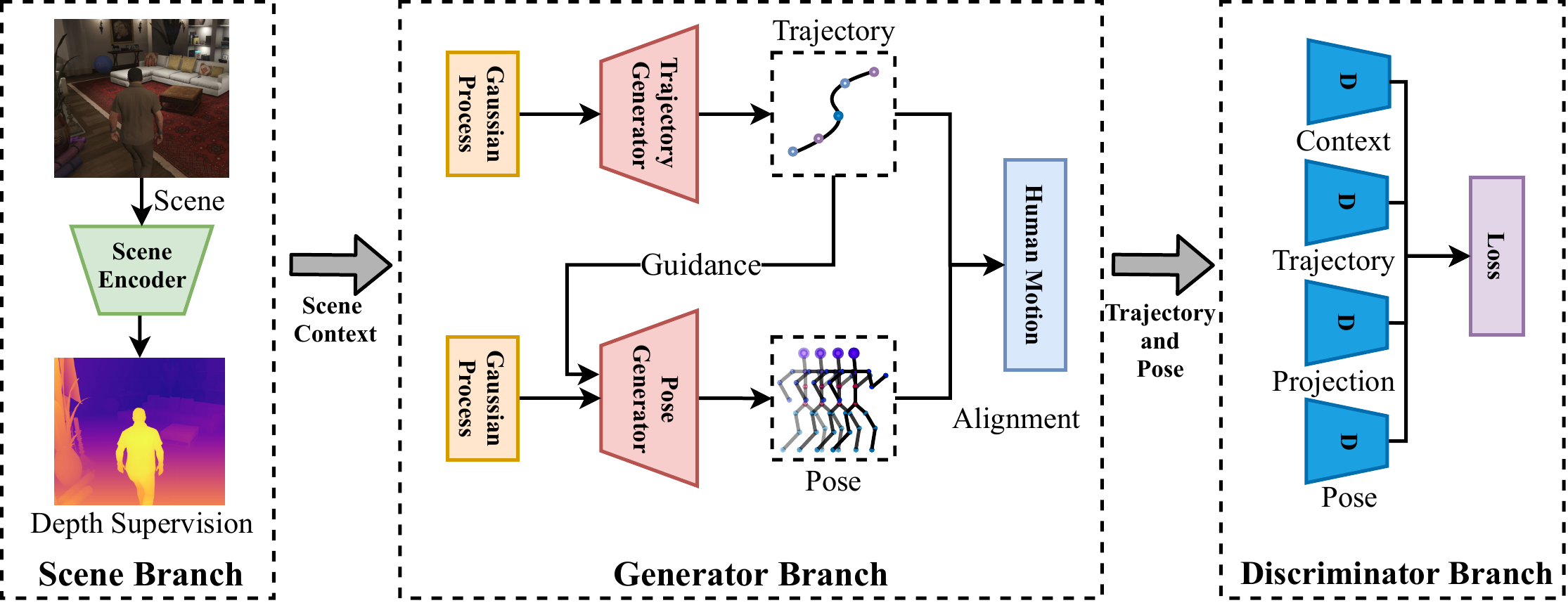}\small
\caption{\textbf{Overview of our framework}.
There are three components in our framework. The first component is the scene branch, which extracts the geometry-aware feature of the given scene and sends it to the generative network as the condition.
The second branch is our motion generator. We first learn the distribution of trajectory, and the pose distribution is learned under the guidance of sampled trajectories. The human motion can be synthesized by aligning different sampled trajectories as well as poses. At last, the discriminator branch is introduced into this framework for the synthesized motions, and the detailed structure of this branch is demonstrated in Figure~\ref{fig:discriminator}.
}
\label{fig:method}
\vspace{-0.5cm}
\end{figure*}
\paragraph{Pose Generation}
Recently, rather than capturing human poses directly~\cite{rong2019delving, rong2020chasing, rong2020frankmocap, wang2020motion}, lots of work begin to focus on pose sequence generation. 
HP-GAN~\cite{barsoum2018hp} combines the Seq2Seq model to the GAN framework for motion generation. 
Cai \emph{et al.}~\cite{cai2018deep} propose a Two-Stage GAN to generate the spatial and temporal information respectively for pose generation.
PSGAN~\cite{yang2018pose} takes the initial pose as input and action label as the condition to generate pose sequence for video generation. 
CSGN~\cite{yan2019convolutional} formulates both generator and discriminator as graph convolution and generates pose sequence from noise sequence directly.
Action2Motion~\cite{chuan2020action2motion} generates human pose sequences with a CVAE model for the given action.
 However, all these methods neglect the role of scene context in motion synthesis, and our framework is the first one for this task, as far as our knowledge.

\paragraph{Pose Prediction}
Pose prediction is also another important task to understand human behaviors.
For given continuous pose sequences, these models can predict the future human motion at a few time steps.
\emph{Encoder-Recurrent-Decoder} (ERD)~\cite{fragkiadaki2015recurrent} incorporates encoder and decoder models before and after the recurrent units for motion prediction. 
Based on the Seq2Seq~\cite{sutskever2014sequence} model, Martinez \emph{et al.}~\cite{martinez2017human} predicts the velocities rather than the positions of joints for motion prediction.
Ac-Lstm~\cite{li2017auto} enhances the capability of LSTM by training the mixture of synthesized frames and observed frames.
Graph convolution network (GCN) is also widely used in motion prediction in recent advances~\cite{Cui_2020_CVPR, Li_2020_CVPR,mao2019learning}.
These methods model dynamic spatial and temporal relationship from the obvious frames to the future frames. 
Recently, more researchers focus on human motion prediction under 2D or 3D scene context~\cite{liu2020group,qi2019amodal, qi2018sequential,qi2020pointins,xing2020malleable, yu2018bisenet}.
Cao \emph{et al.}~\cite{cao2020long} propose a three-stage motion prediction method which can predict different human motions under different destinations.
However, our method is significantly different against previous motion prediction methods.
Our fully generative network can directly learn the distribution of human in the scene and synthesize diverse human motions, as shown in Figure~\ref{fig:teaser}, rather than the deterministic prediction.

%% file: Sections/section_method.tex
\section{Scene-aware Generative Network}~\label{Method}

At first, we formally define the problem of human motion synthesis given the scene context.
The human motion is represented as $X = (R, P)$,
where $R$ stands for the trajectory in the scene, and $P$ stands for the pose sequence.
Inspired by previous methods~\cite{cao2020long,li2019putting,PSI:2019}, we represent the scene as an RGB image.
Besides the RGB image, we also provide the initial pose following previous methods~\cite{cao2020long,chuan2020action2motion},
and they together constitute the input condition $S$.
Without loss of generality, we fix the start point of $R$ at the image center,
since we can always crop the image around the start point.
Based on these notations,
human motion synthesis under the given scene context thus can be described as sampling a valid $X$ from a conditional distribution:
\begin{equation}
    X \sim p((R, P)| S).
\end{equation}

Instead of learning a deterministic mapping from a given $S$ to some $X$,
we propose a fully generative framework based on generative adversarial networks~\cite{goodfellow2014generative}
to directly model the distribution $p((R, P)|S)$,
so that the diversity of $X$ under the given $S$ can be fully captured.
Moreover,
inspired by the observation that humans usually subconsciously plan a rough trajectory in mind before moving,
we divide the conditional distribution $p((R, P)|S)$ into two independent distributions as:
\begin{equation}
    p((R, P)|S) = p(R|S) \cdot p(P|R, S),
\end{equation}
and further organize our framework in two stages consisting of two different GANs.
These two GANs respectively capture $p(R|S)$ and $p(P|R, S)$
and are jointly trained in an end-to-end manner.
Such a decomposition not only significantly reduces the computational complexity
but also provides the flexibility to add trajectory- or pose-specific constraints to the proposed framework.
Specifically,
we include two extra discriminators, namely a projection discriminator and a context discriminator
to emphasize the consideration of scene context when modeling $p(R|S)$.
An overview of our final two-stage generative framework with scene-aware constraints is included in Fig.~\ref{fig:method}.
Below we briefly introduce these components separately,
and include their detailed architectures in the supplemental materials.

\subsection{Scene Encoder}
\label{sec:scene_encoder}
Given a scene in the form of an RGB image,
a good scene encoder $E_\mathrm{scene}$ should reflect both the visual scene semantics and the scene structure
in its extracted scene feature $\vf_\mathrm{scene}$,
as humans are likely to move following the scene structure
and perform actions that are semantically consistent with the scene,
\eg~going upstairs when there is a stairway in the scene.
To capture visual semantics of the scene,
we deploy a ResNet-18~\cite{he2016deep} pre-trained on ImageNet \cite{deng2009imagenet} as the scene encoder.
More importantly,
depth estimation is added as an auxiliary task for $E_\mathrm{scene}$ when training it with the whole framework jointly.
To estimate the depth map from the given scene image,
$E_\mathrm{scene}$ is required to include the scene structure in the extracted feature $\vf_\mathrm{scene}$.
In practice, depth estimation is achieved via an inverse huber loss $\cL_\mathrm{scene}$,
which is commonly adopted in recent works for monocular depth estimation \cite{laina2016deeper}.

\begin{figure}[t]
\centering
\includegraphics[width=0.9\linewidth]{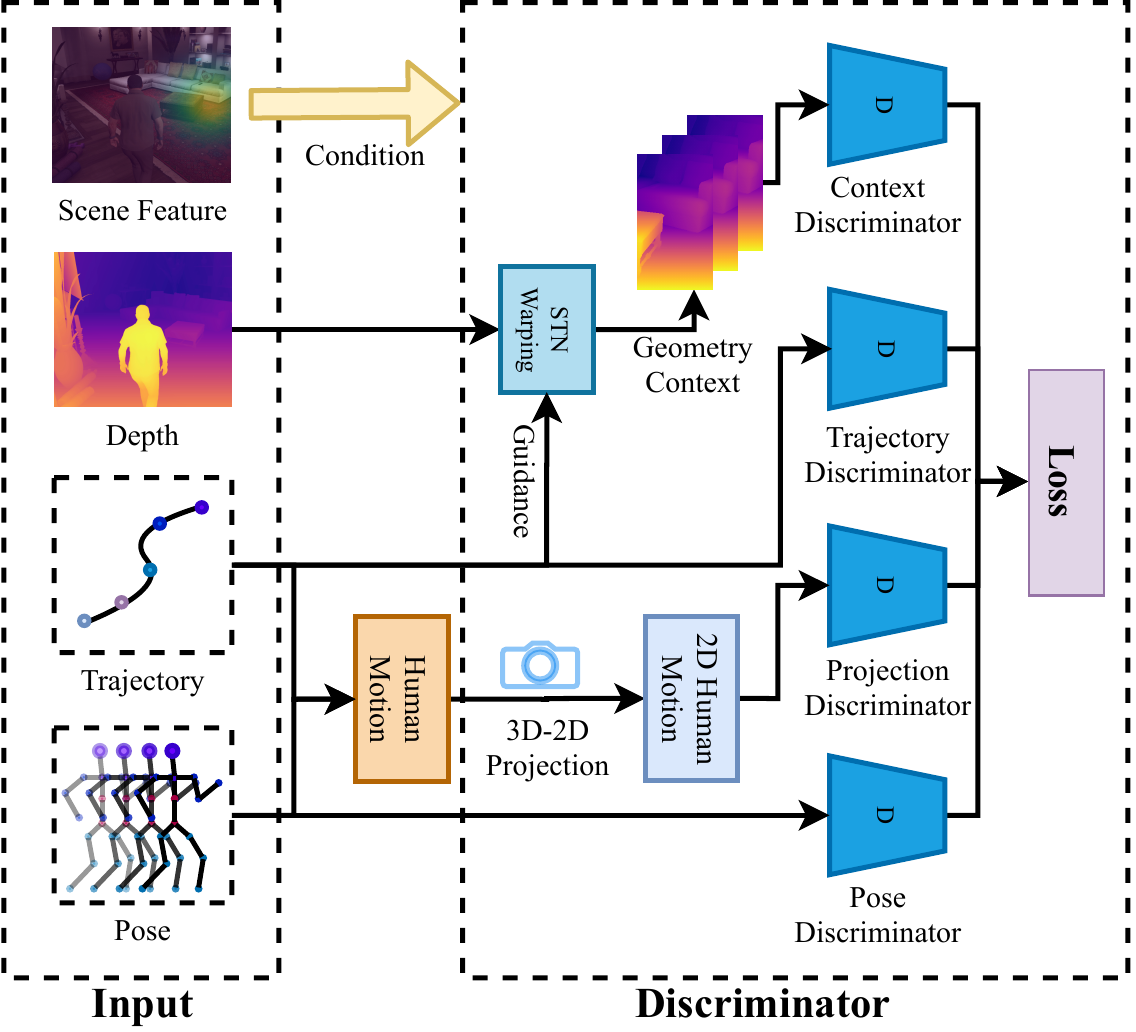}
\caption{\textbf{Overview of our discriminator branch}. This discriminator branch contains four discriminators. The trajectory and pose discriminator help the synthesized human motions to be smooth and continuous. The projection and context discriminator encourage the human motions following the physical structure of the scene.
}\vspace{-0.5cm}

\label{fig:discriminator}
\end{figure}
\subsection{Generator Branch}~\label{G}
As mentioned, the proposed generative framework naturally decomposes $p((R, P)|S)$ into two independent distributions, namely $p(R|S)$ and $p(P|R, S)$.
Subsequently,
a trajectory generator is used to model $p(R|S)$ and synthesize trajectories under the condition $S$.

Once a trajectory $R$ is synthesized,
at each time step $R$ provides the location in the scene, as well as the moving speed and orientation.
Since all this information served as strong priors for the distribution of pose sequences,
a pose generator is thus adopted to learn $p(P|R, S)$ conditioned on both $S$ and $R$.
Finally, we append the trajectory $R$ to the root joint of the pose sequence $P$ to form our final output.

\subsubsection{Trajectory Generator}~\label{Path Generator}
The trajectory generator $G_\mathrm{traj}$ represents a trajectory $R$ of length $T_R$
as a sequence of velocities $V=(\vv_0, \vv_2, ..., \vv_{T_R-1})$
so that the location $R_t$ at time step $t$ is obtained via $R_t = \sum_{i = 0}^{t - 1} \vv_i$.
To learn $V$, $G_\mathrm{traj}$ follows a process where it gradually increases the sequence length to $T$,
based on the scene feature $\vf_\mathrm{scene}$ and the initial pose.
Such a coarse-to-fine process is shown to effectively reduce the learning complexity.
Specifically,
we at first sample a sequence of latent vectors $Z=(\vz_0, \vz_1, ..., \vz_{T^{(0)}-1})$
where its length $T^{(0)}$ is smaller than the final length $T_R$,
and each $\vz$ is a $d_\vz$-dimensional vector.
Since trajectories should be smooth,
the sequence $Z$ is set to follow a Gaussian Process,
where $c$-th component of $[\vz_0, ..., \vz_{T^{(0)}-1}]$ satisfies $[\vz_0^{(c)}, ..., \vz^{(c)}_{T^{(0)}-1}] \sim GP(\vzero, \kappa)$ with
$\kappa(t_1, t_2) = \exp(-\frac{|t_1 - t_2|}{2 \sigma_c^2})$.
We change the value of $\sigma_c$ for each $c$ to encourage them capture different temporal correlations.
After $Z$ is sampled,
$G_\mathrm{traj}$ will use its first block $G^{(1)}_\mathrm{traj}$ consisting of deconvolution layers
to upsample $Z$ into a feature sequence $F^{(1)}$ of length $T^{(1)}$ where $T^{(1)} = 2 T^{(0)}$.
This upsampling step will be repeated $K$ times
so that $F^{(K)} = G_\mathrm{traj}^{(K)} \circ G_\mathrm{traj}^{(K - 1)} \circ ... \circ G_\mathrm{traj}^{(1)} (Z)$
and $T^{(K)} = T_R = 2^K T^{(0)}$.
While each feature in $F^{(K)}$ is a scalar,
the velocity sequence $V$ is obtained by $V = \alpha \tanh(F^{(K)})$ conducted at each time step.
The hyperparameter $\alpha$ is used to control the average moving speed.
It's worth noting the velocity sequence is modeled in the 3D camera coordinate system
rather than the 2D image coordinate system
to avoid potential issues raised by the 3D-2D projection.

\subsubsection{Pose Generator}~\label{Pose Generator}
Conditioned the scene feature $\vf_\mathrm{scene}$, the initial pose,
and the synthesized trajectory $R$ of length $T_R$,
the pose generator $G_\mathrm{pose}$ will output a pose sequence $P$ consisting of $T_P=T_R$ poses
each of which contains $J$ joints.
$G_\mathrm{pose}$ follows a similar structure with the trajectory generator $G_\mathrm{traj}$.
Starting from a sequence of latent vectors $Z^\prime$ that follows another Gaussian Process,
$G_\mathrm{pose}$ gradually upsamples $Z^\prime$ into the pose sequence $P$
via blocks of graph-upsampling layers, as in \cite{yan2019convolutional}.
It's worth noting that the pose sequence is also represented in the 3D camera coordinate system,
and all poses in the synthesized pose sequence are center subtracted
as their movements in the scene will be controlled by the synthesized trajectory $R$ as $X = P + R$.

\subsection{Discriminator Branch}~\label{D}
\begin{figure}[t]
\footnotesize
\centering
\renewcommand{\tabcolsep}{1pt} 
\begin{center}
\begin{tabular}{cc}
\includegraphics[width=0.65\linewidth]{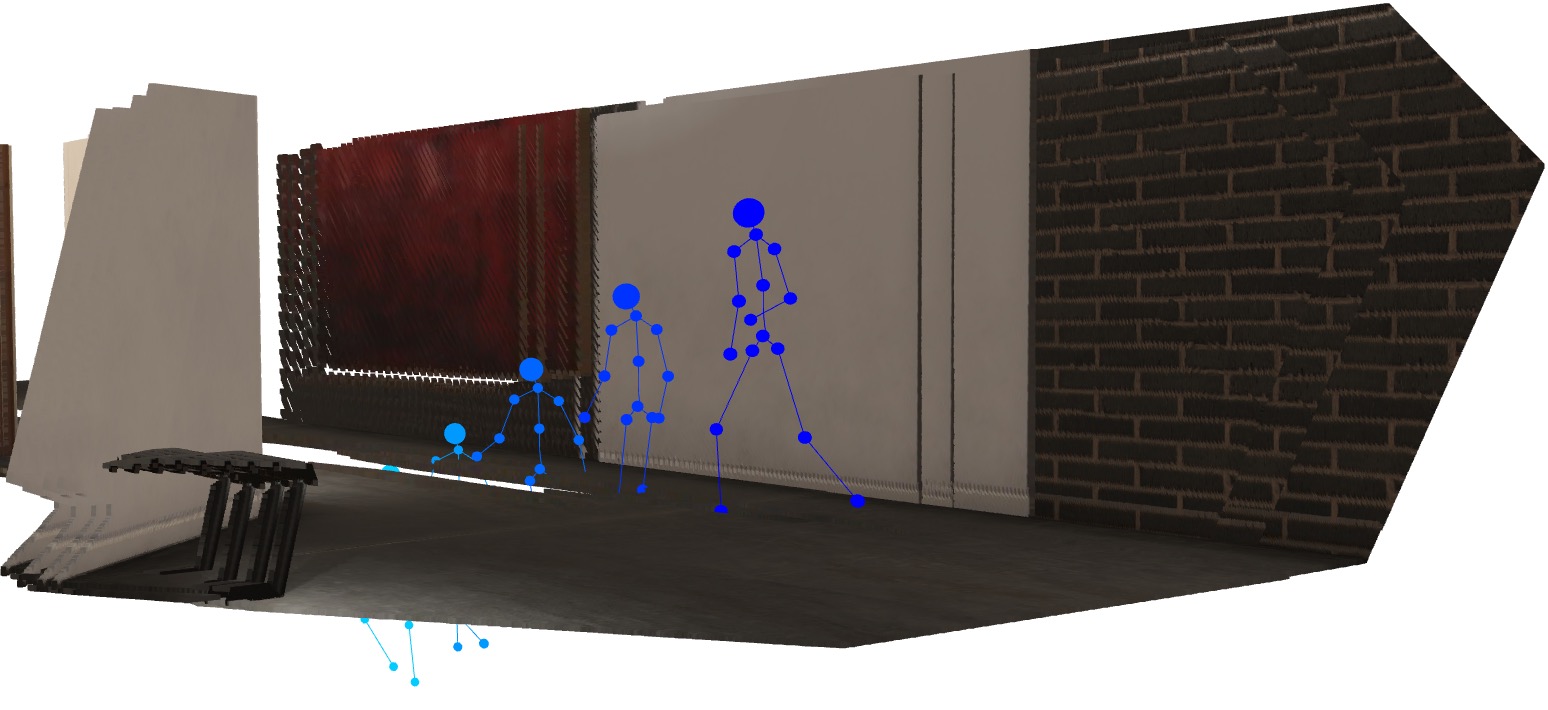} &
\includegraphics[width=0.3\linewidth]{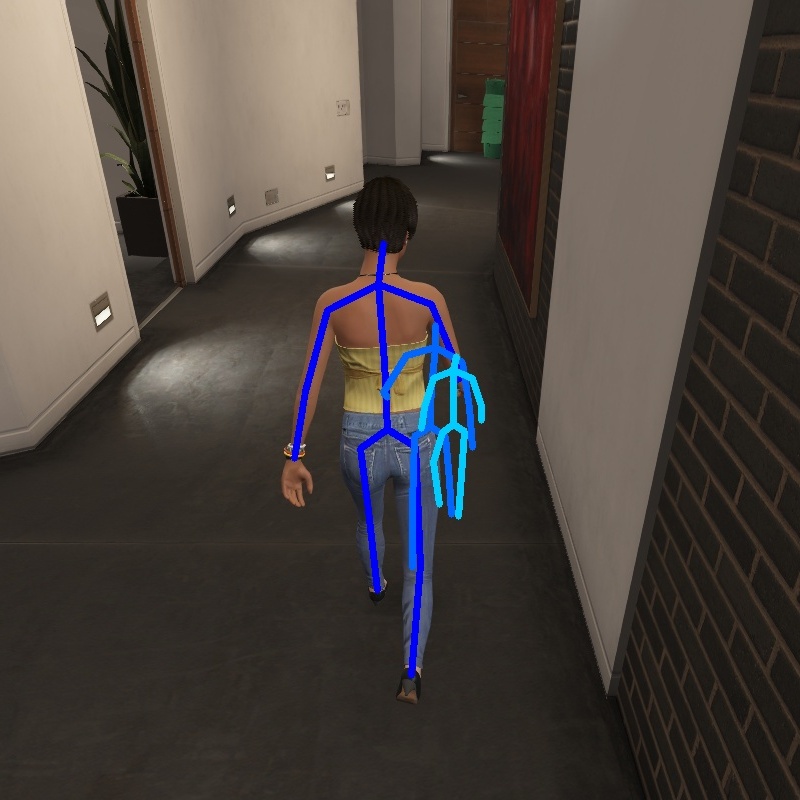} \\
\includegraphics[width=0.65\linewidth]{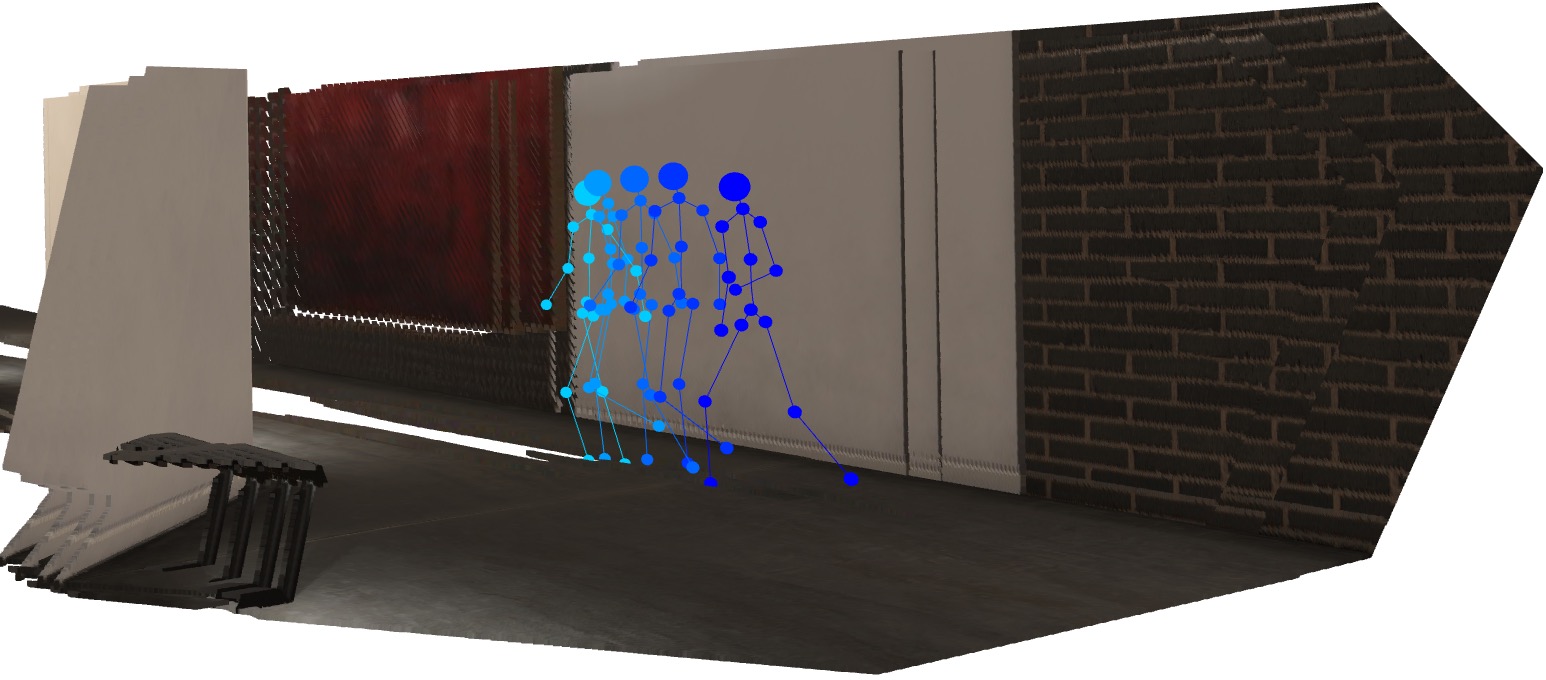} &
\includegraphics[width=0.3\linewidth]{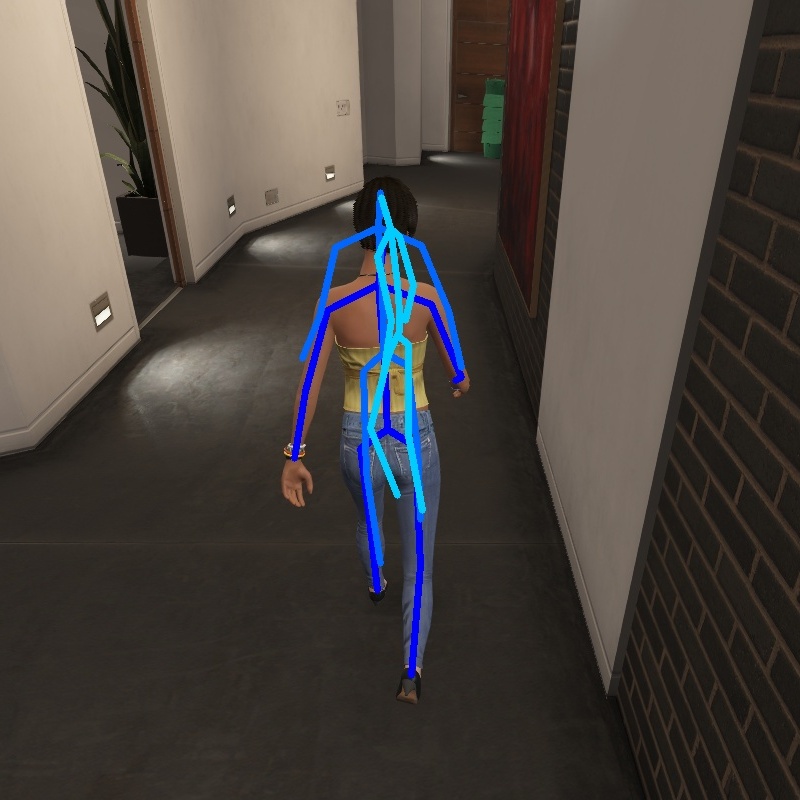} \\
(a) 3D Human Motion &
(b) 2D Human Motion \\

\end{tabular}
\end{center}
\caption{\textbf{Visualization of 3D human motion and the projection of this motion in 2D space}. The results in the first row is sampled from the method without projection discriminator and the second is supervised by this discriminator.
With the scale changes of human motions in the 2D scene, the geometry compatibility of them can be easily judged.}\vspace{-0.4cm}

\label{fig:motion_projection}
\end{figure}

While both the trajectory generator $G_\mathrm{traj}$ and the pose generator $G_\mathrm{pose}$ have
a corresponding discriminator, namely $D_\mathrm{traj}$ and $D_\mathrm{pose}$,
to distinguish synthesized trajectories and pose sequences from real ones,
relying on only the scene feature $\vf_\mathrm{scene}$ to reflect the scene structure is insufficient as shown in Fig.~\ref{fig:motion_projection},
where the synthesized trajectory and pose sequence looks natural when viewed in isolation,
but lead to some inconsistency when combined with the scene.
Therefore, to further enhance the compatibility between synthesized human motions and the given scene,
we propose two additional discriminators,
namely the projection discriminator $D_\mathrm{proj}$ and the context discriminator $D_\mathrm{context}$,
that respectively focus on global structural constraints such as walls, the floor or the ceiling,
and local structural constraints such as chairs and tables.

\subsubsection{Projection Discriminator}~\label{projection discriminator}
The projection discriminator $D_\mathrm{proj}$ is used to
enhance the compatibility between synthesized human motion $X$ and the global scene structure,
so that $X$ will not result in the human goes through the wall, collides into the floor, or floats in the air.
One challenge to solve when applying $D_\mathrm{proj}$ is that
the motion $X$, either real or fake, is represented in the 3D camera coordinate system,
while the scene is described using a 2D image,
the mismatch between 3D and 2D thus may lead to an inferior discriminator.
Fortunately, $D_\mathrm{proj}$ will only be deployed during training,
and the camera matrix of each scene image is often provided by existing datasets.
Therefore,
$D_\mathrm{proj}$ will project the input human motion $X$ into the image coordinate system, acquiring its 2D counterpart $X^\prime$.
As shown in Figure~\ref{fig:motion_projection},
the projected human motion $X^\prime$ can be effectively represented
by the changes of 2D coordinates and the scale of human in the scene,
which are relatively easier for the discriminator to make judgments.
Since $X^\prime$ is a trajectory-aligned 2D pose sequence,
$D_\mathrm{proj}$ is built upon graph convolution layers with graph downsampling,
with a final global average pooling layer to aggregate information across all time steps.

\begin{figure}[t]
\centering
\includegraphics[width=0.8\linewidth]{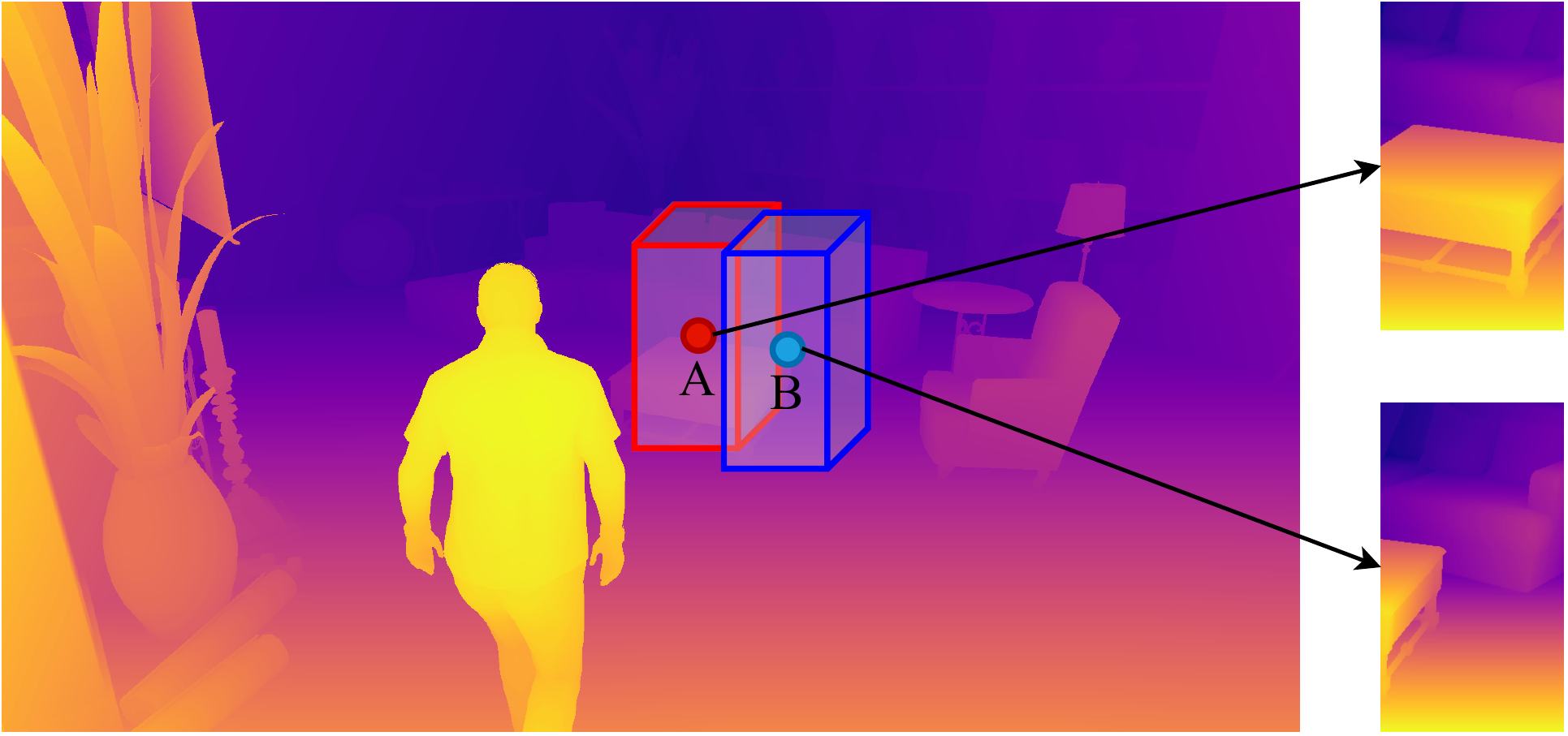}
\caption{\textbf{Visualization of the local context for context discriminator}. {\color{red} Point A} is the collision position to the desk and the {\color{blue} point B} is the compatible position of the scene for human motions. The geometry context of these two positions are significantly different.
}\vspace{-0.3cm}

\label{fig:context}
\end{figure}
\subsubsection{Context Discriminator}~\label{context discriminator}
Besides the projection discriminator $D_\mathrm{proj}$,
we also include a context discriminator $D_\mathrm{context}$ to
encourage the compatibility between the synthesized human motion $X$ and the local scene structure,
which mainly refers to the constraint that a human should avoid hitting objects in the scene when moving.
Such a constraint requires $D_\mathrm{context}$ to have
a detailed understanding of the local scene context around $X$ at each time step.
On the other hand,
as shown in Figure~\ref{fig:context},
when a human motion $X$ is incompatible with the local scene structure,
the relative depth between it and the scene contains meaningful patterns.
Inspired by this observation,
$D_\mathrm{context}$ utilizes local relative depth crops as its main source of judgments.
Specifically,
$D_\mathrm{context}$ will also project the input human motion $X$ into its 2D counterpart $X^\prime$,
followed by cropping the scene depth map around $X^\prime$ at each time step.
The crop at $t$-th time step has the shape $(\frac{H_c}{d_t}, \frac{W_c}{d_t})$,
where $d_t$ is the corresponding depth value of $X^\prime$ at that time,
and $(H_c, W_c)$ is the pre-defined crop size (in practice, we set it to be $\frac{1}{4}$ of the image size.).
All these local depth crops will subtract their corresponding depth values of $X^\prime$
and resize to $(H_c, W_c)$,
forming the local relative depth crop sequence.
It's worth noting the scene depth map is estimated from the scene feature $\vf_\mathrm{scene}$ as discussed in Section~\ref{sec:scene_encoder},
and the Spatial Transformer Network (STN)~\cite{jaderberg2015spatial} is used as a differentiable cropping function.

%% file: Sections/section_experiments.tex
\section{Experiments}

\subsection{Dataset}
In this paper, we mainly evaluate the proposed method on two public datasets, which contain both 3D human motions and geometry information of different scenes.

\textbf{GTA-IM:}
GTA-IM~\cite{cao2020long} is a recent dataset based on the virtual GTA environment.
This dataset contains long-term and diverse human motions in different indoor and outdoor scenes, as well as sufficient information of the scene, including depth maps and instance labels.
We select 70 sequences in 6 scenes as the training set and 30 sequences in 3 scenes as the testing set.
Then we sampled sub-sequences by a sliding window with a fixed length of 65 frames and a step of 5 frames, where the begin frame is as the condition for motion synthesis.
Two scenes in the testing set are also shared with the training set, and one scene is exclusive.
Although these two sets contain the sequences in the same scene, they are synthesized in different camera viewpoints and the initial pose for generation is also different.
Therefore, testing our method on same scenes is non-trivial on this dataset.
Similar to \cite{cao2020long}, we only choose 19 key joints from 98 human joints provided by this dataset for human motion synthesis.

\textbf{PROX:}
Proximal Relationships with Object eXclusion (PROX)~\cite{PROX:2019} is captured by the Kinect.
This dataset contains 12 different 3D scenes and provides 3D human model sequences.
To evaluate our method in the real-scene, we split 10 scenes as the training set and 2 scenes as the testing set.
To improve the quality of synthesized motions, 
we first train our model on GTA-IM and the fine-tune it on this dataset.
We select 16 joints from the SMPLX~\cite{SMPL-X:2019} model and all these joints are contained in GTA-IM.

\subsection{Implementation Details}
We adopt Adam~\cite{kingma2014adam} algorithm under (0.5, 0.999) betas and 0.0002 learning rate as optimizer for all the experiments on both GTA-IM and PROX datasets.
Following~\cite{yan2019convolutional}, our model is supervised by the WGAN-GP~\cite{gulrajani2017improved} framework.
The gradient penalty is 10 and the gradient regression target is 0.1.
We train our model 150 epochs and generate 64 frame pose sequences on both GTA-IM and PROX dataset.

\subsection{Metric}
Since the previous works did not propose metrics to measure the quality of scene-aware action generation,
we introduce different metrics to evaluate the different characteristics of generated pose sequences, such as motion, pose, and compatibility with the given scene.

\textbf{Motion FID:}
In CSGN~\cite{yan2019convolutional}, the well-known metric \emph{Frm\'{e}chet Inception Distance} (FID) has been extended to pose generation task.
Because explicit action labels are not available in both GTA-IM and PROX, we change the encoder model from ST-GCN~\cite{yan2018spatial} model, which is used to compute FID score in~\cite{yan2019convolutional}, to \textbf{ERD}~\cite{fragkiadaki2015recurrent}, which is a well-known LSTM based method for motion prediction.
Therefore, this FID score is extend to measure the continuity and rationality of generated pose sequences by the action agnostic features.
To well measure the quality of the synthesized motion in different lengths, firstly, we conduct different ERD models, who are trained based on the human motions with 16, 32, and 64 frames. Then we clip the synthesized $T$ frames human motion to multiple short human motions.
The clip lengths of these sequences are 2, 4, 8, 16, 32, and 64.
The average value of the fid score is calculated based on the composition of these ERD models and the clip length as our Motion FID.
Besides, we define the mean fid for 2 and 4 length as short-term value, 8 and 16 as middle-term value, and 32 and 64 as long-term value.

\begin{figure}[t]
\centering
\includegraphics[width=0.8\linewidth]{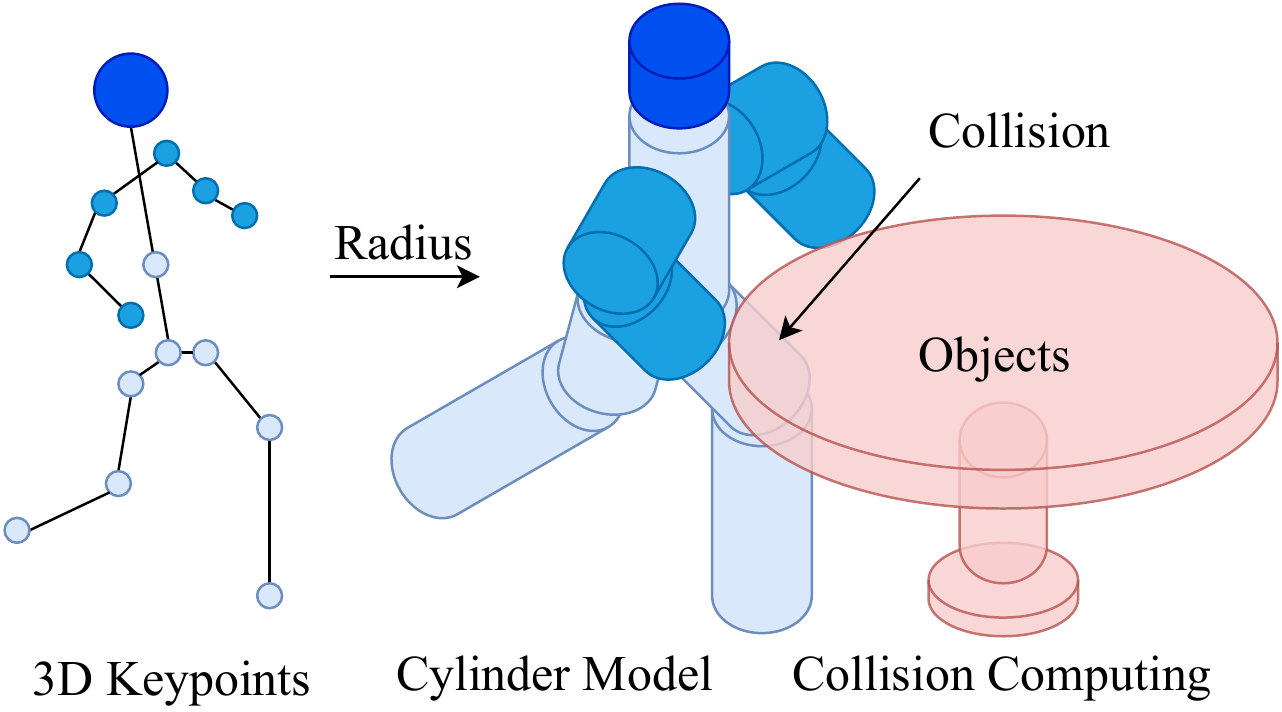}
\caption{\textbf{Collision between human cylinder model and objects}. Pre-defined cylinder radius is conducted to the bones of the synthesized 3D human motions. We define the collision by computing the point number of 3D objects in this cylinder model.
}\vspace{-0.5cm}
\label{fig:collision}\end{figure}

\textbf{Non-collision Ratio:}
Besides the motion FID, we also propose the non-collision ratio as metric for physical compatibility between human motions and the given scene.
Firstly, we conduct cylinder model to synthesized skeleton sequences by pre-defined radius, as shown in ~\ref{fig:collision}.
Then, the human-scene collision area can be represented as the intersection points between these cylindricalization pose sequences and the point cloud of given scenes, which are projected from the 2D depth map by camera parameters.
Under this representation, the collision motion can be defined as motions with intersection points more than the pre-defined threshold \textbf{\emph{t}}. 
Thus, our non-collision ratio can be calculated by the ratio between the number of human motions without human-scene collision and all sampled motions.
In this work, we discuss non-collision ratio with different thresholds \textbf{\emph{t}}: 40, 60, 80, and 100 and the radius of cylinder model with different \textbf{\emph{r}}: 30, 45, and 60\emph{mm}. We randomly sample 10000 different frames in GTA-IM and 2000 frames in PROX for ablation studies. 

\textbf{User Study:}
At last, we conduct user study to further evaluate the quality of the synthesized human motions,
especially the rationality of interaction between generated human motions and the scene.
In this user study, 
we let the users focus on the two kinds of compatibilities:
one between the human motions and the given scene, and one between the trajectory and pose sequence.
We named these two scores as scene compatibility and motion compatibility, respectively.
For this user study, we mixed the human motions from the ground-truth, 
the origin CSGN, our method without geometry discriminators, 
and the proposed method in the same scene,
and sample equal numbers of human motions for each approach.
The score for this user study is between 1 (strongly abnormal) and 5(strongly compatible).
And we will illustrate the mean and standard deviation of the scores from users.

\subsection{Ablation Studies}~\label{Ablation}

\begin{figure}[t]
\centering
\includegraphics[width=0.95\linewidth]{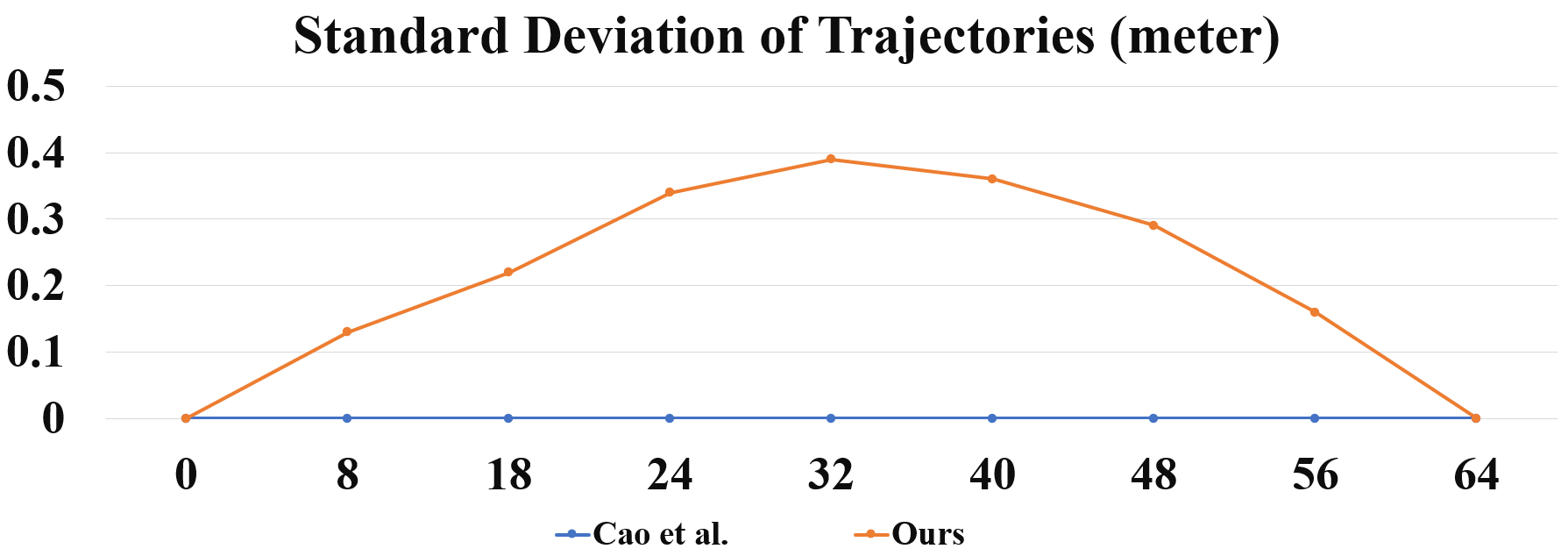}
\caption{Standard deviation over time of 5 synthesized trajectories given the same start and end points for images in GTA-IM, where~\cite{cao2020long} obtains 0 metre since it's deterministic in this case. 
}

\label{fig:divseristy}
\end{figure}

\begin{table}[t]
\centering
\small
\renewcommand{\tabcolsep}{5pt}
\caption{Ablation studies on GTA-IM dataset. 
M: separately generate trajectory and pose; D: depth supervision; P: projection discriminator; 
C: context discriminator.}~\label{Tab:ablation study}
\begin{tabular}{l|p{5pt}p{5pt}p{5pt}p{5pt}|cccc}
\shline
\multicolumn{9}{c}{Motion FID}\\
\hline
Method & M & D & P & C & Short & Mid & Long & Ave $\downarrow$ \\     
\hline
CSGN & & & & & 92.4 & 125.6 & 137.8 &118.6 \\
\hline
\multirow{5}{*}{Ours} & \checkmark & & & & 27.6 & 33.4 &38.6 & 33.1 \\
 & \checkmark & \checkmark & & &26.8 & 32.2 &37.5 & 32.1 \\
 & \checkmark & \checkmark & & \checkmark &25.9 & 31.0 &35.6 & 30.8\\
 & \checkmark & \checkmark & \checkmark & & 23.3 & 28.4 & 34.5 & 28.7 \\
 & \checkmark & \checkmark & \checkmark & \checkmark & 22.4 & 27.6 & 33.7 & 27.6 \\
\hline
\multicolumn{9}{c}{Non-Collision Score}\\
\hline
Method & M & D & P & C & 30mm & 45mm & 60mm & Ave $\uparrow$ \\     
\hline
\multirow{5}{*}{Ours} & \checkmark & & & & 0.931& 0.882 & 0.851 & 0.891 \\
 & \checkmark & \checkmark & & & 0.941& 0.902 & 0.874 & 0.902 \\
 & \checkmark & \checkmark & & \checkmark & 0.948 & 0.919 & 0.904 & 0.923 \\
 & \checkmark & \checkmark & \checkmark & & 0.956 & 0.934 & 0.915 & 0.938 \\
 & \checkmark & \checkmark & \checkmark & \checkmark& 0.962 & 0.943 & 0.924 & 0.943 \\
\shline
\end{tabular}
\label{tab:ablation}
\end{table}

In this section, we will first conduct ablation studies on GTA-IM dataset to demonstrate the effectiveness of our framework in detail.
As shown in Table~\ref{Tab:ablation study}, we demonstrate the Motion FID and Non-Collision Score of the models with different module combinations.
Firstly, with separation synthesis between trajectory and pose, Motion FID decreases significantly (118.6 $\rightarrow$ 33.1), which verifies the effectiveness of our design choices on generator branch.
Then, the Motion FID is mitigated with discriminators for physical compatibility, and the Non-Collision score is improved from 0.902 to 0.943.
The influence of depth supervision is also demonstrated in this table. The Motion FID is decreased from 33.1 to 32.1, as well as the Non-Collision score is improved from 0.891 to 0.902. 
Moreover, as the ablation studies on GTA-IM, we demonstrate the Motion FID and Non-collision score on PROX in Table~\ref{tab:fid-prox} and Table~\ref{tab:collision}, respectively. Our geometry-aware discriminators improve this score from 0.742 to 0.792, and the ability of them to synthesize physical compatibility motions is verified in captured real scenes. Besides, the Motion FID is mitigated from 161.1 to 54.6 based on our framework.

\begin{table}[t]
\centering
\small
\renewcommand{\tabcolsep}{9pt}
\caption{
Comparison of Motion FID on \textbf{GTA-IM}. M: separately generate trajectory and pose; G: geometry-aware discriminator.}
\label{tab:fid-gta}
\begin{tabular}{l |c c c c}
\shline
 Method & Short & Mid & Long & Ave $\downarrow$  \\
\hline
Two-Stage~\cite{cai2018deep} & 172.8 & 196.5 & 210.4  & 193.2\\
HP-GAN~\cite{barsoum2018hp} & 156.6& 176.2 & 190.4& 174.4\\
CSGN~\cite{yan2019convolutional}& 92.4& 125.6 & 137.8 & 118.6\\
\hline
CSGN+ \emph{M}  & 26.8& 32.2 & 37.5& 32.1\\
CSGN+ \emph{M} + \emph{G} & 22.4& 27.6 & 33.7 & \textbf{27.6} \\
\shline
\end{tabular}
\end{table}

\begin{table}[t]
\centering
\small
\renewcommand{\tabcolsep}{9pt}
\caption{
Comparison of Motion FID on on \textbf{PROX}. M: separately generate trajectory and pose; G: geometry-aware discriminator.}
\label{tab:fid-prox}
\begin{tabular}{l |c c c c}
\shline
 Method & Short & Mid & Long & Ave $\downarrow$  \\
\hline
Two-Stage~\cite{cai2018deep} & 261.5 & 276.8 & 292.4  & 276.9\\
HP-GAN~\cite{barsoum2018hp} & 214.7& 225.3 & 251.7& 230.6\\
CSGN~\cite{yan2019convolutional}& 141.5& 160.3 & 183.2 & 161.6\\
\hline
CSGN+ \emph{M}  & 55.3& 58.4 & 62.5& 58.7\\
CSGN+ \emph{M} + \emph{G} & 52.4& 54.5 & 56.9 & \textbf{54.6} \\
\shline
\end{tabular}
\end{table}
To further evaluate the effectiveness of proposed method, especially the design choices of our generator and discriminator, we compare our framework against different state-of-the-art methods in Table~\ref{tab:fid-gta} and Table~\ref{tab:fid-prox}.
All the methods in these tables are condition on the same scene context.
The significant mitigation on Motion FID demonstrates the ability of our framework on synthesizing human motions in the scene.
Besides these motion synthesis framework, we compare our framework against~\cite{cao2020long} in Figure~\ref{fig:divseristy}.
Under the same start and end points, this motion standard deviation demonstrate the diversity of sampled human motions rather than deterministic prediction.

At last, we also demonstrate the results of the user study on GTA-IM and PROX in Table~\ref{tab:user study}. It is not surprising that the ground-truth human motions achieve the highest score on both datasets. For theses datasets, we observe that the geometry discriminators can significantly improves the performance for learning compatible human motions with the scene. And the two-stage human motion synthesis framework can also help the generator to synthesize smooth and continuous human motions.

\begin{table}[t]
\centering
\small
\renewcommand{\tabcolsep}{10pt}
\caption{
Collision score of the method with and without the supervision of geometry discriminator on PROX.P: projection discriminator; C: context discriminator}
\label{tab:collision}
\begin{tabular}{c c | c c c c}
\shline
 P & C & 30\emph{mm} & 45\emph{mm} & 60\emph{mm} & Ave $\uparrow$  \\
\hline
           &            & 0.793 & 0.731 & 0.692 & 0.742 \\
\checkmark &            & 0.813 & 0.764 & 0.733 & 0.770 \\ 
\hline
\checkmark & \checkmark & 0.831 & 0.782 & 0.753 & \textbf{0.792}\\
\shline
\end{tabular}
\end{table}

\begin{figure*}[t]
\centering
\includegraphics[width=0.88\linewidth]{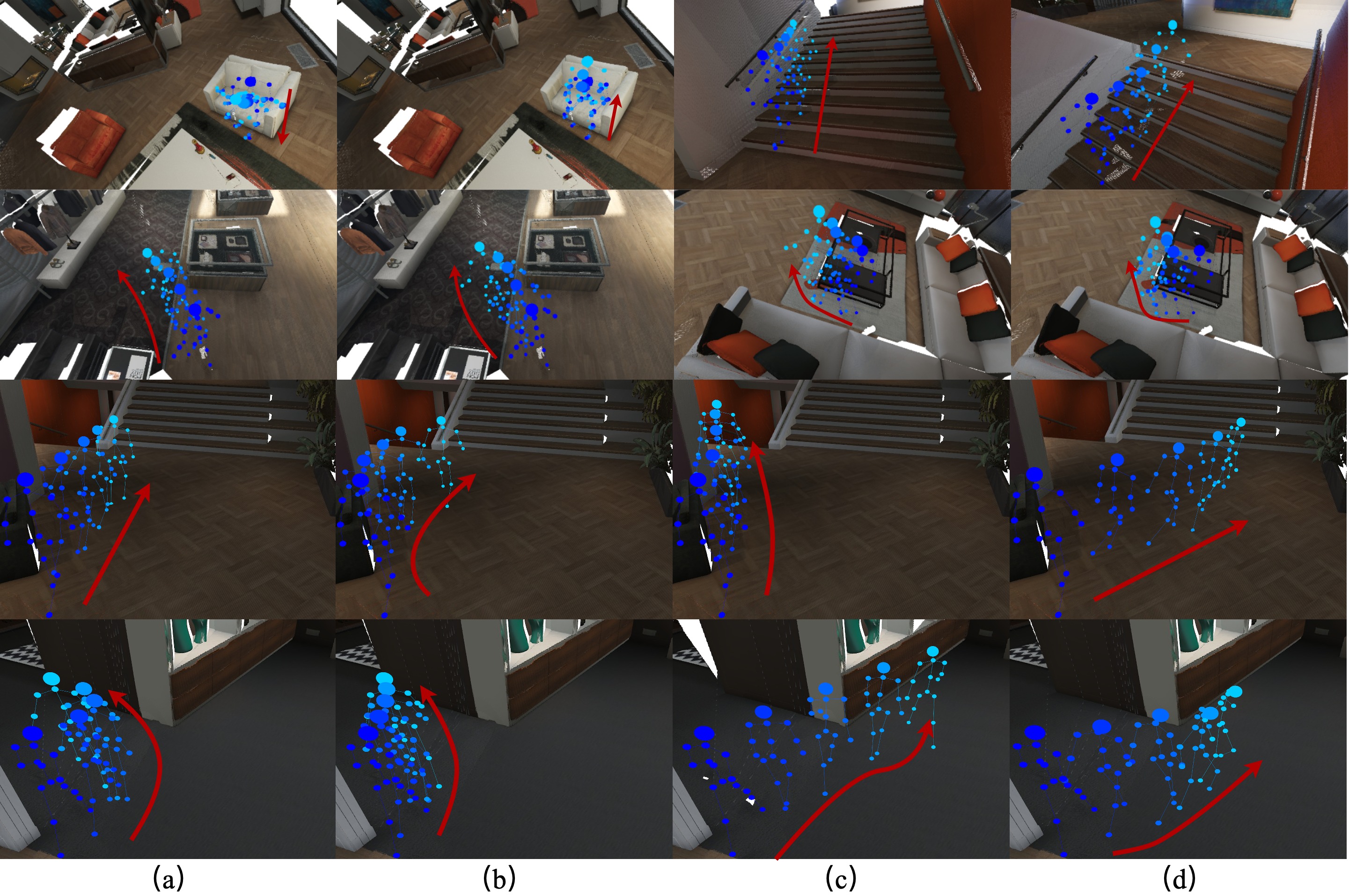}
\caption{\textbf{Visualization on GTA-IM}. All the results are sampled from the synthesized 64 frame sequences. Our framework can synthesize diverse and compatible human motions in different scenes.}\vspace{-0.3cm}
\label{fig:gta_results}

\end{figure*}

\begin{figure}
\footnotesize
\centering
\renewcommand{\tabcolsep}{1.2pt} 
\begin{center}
\begin{tabular}{cc}
\includegraphics[width=0.42\linewidth]{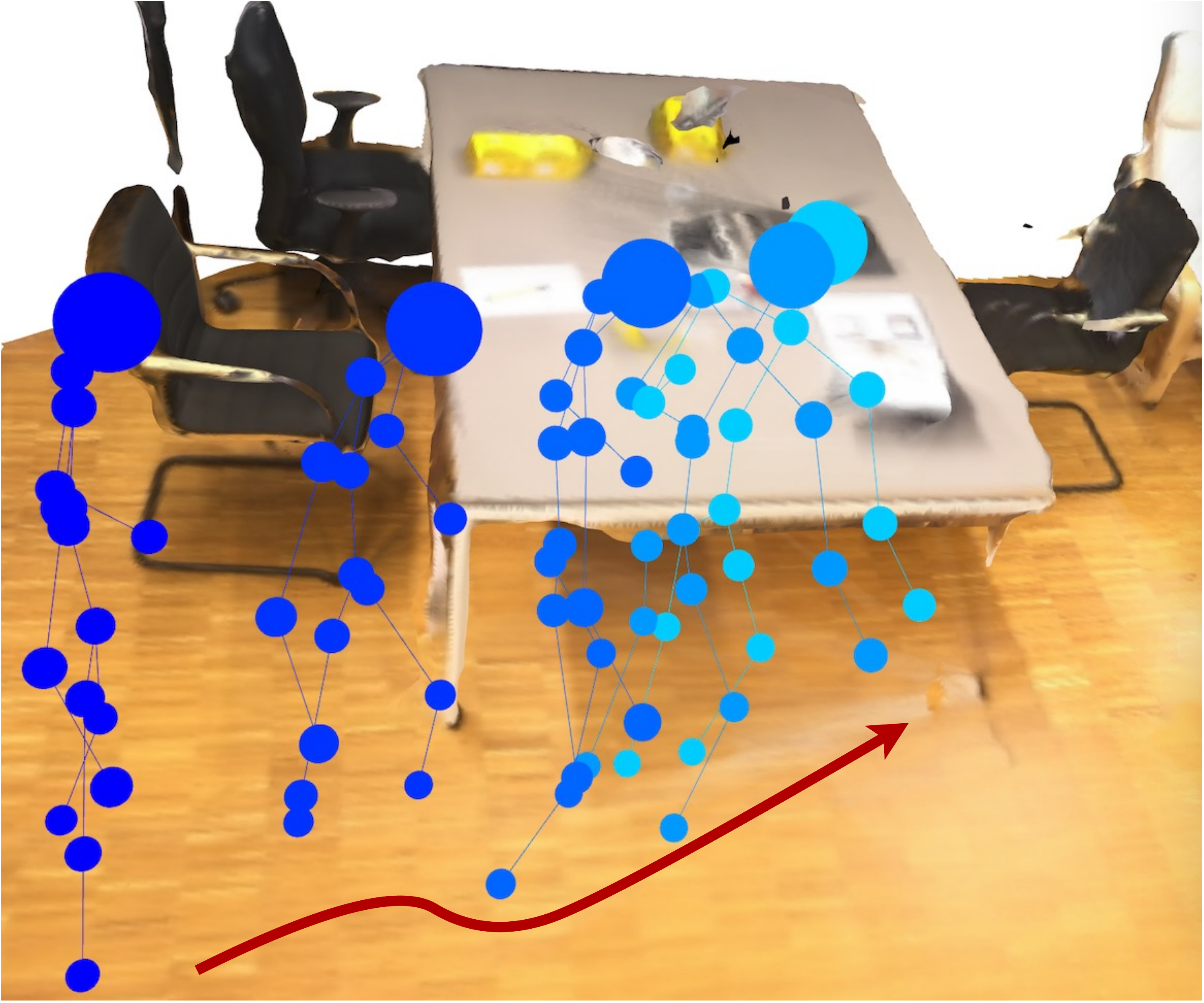} &
\includegraphics[width=0.42\linewidth]{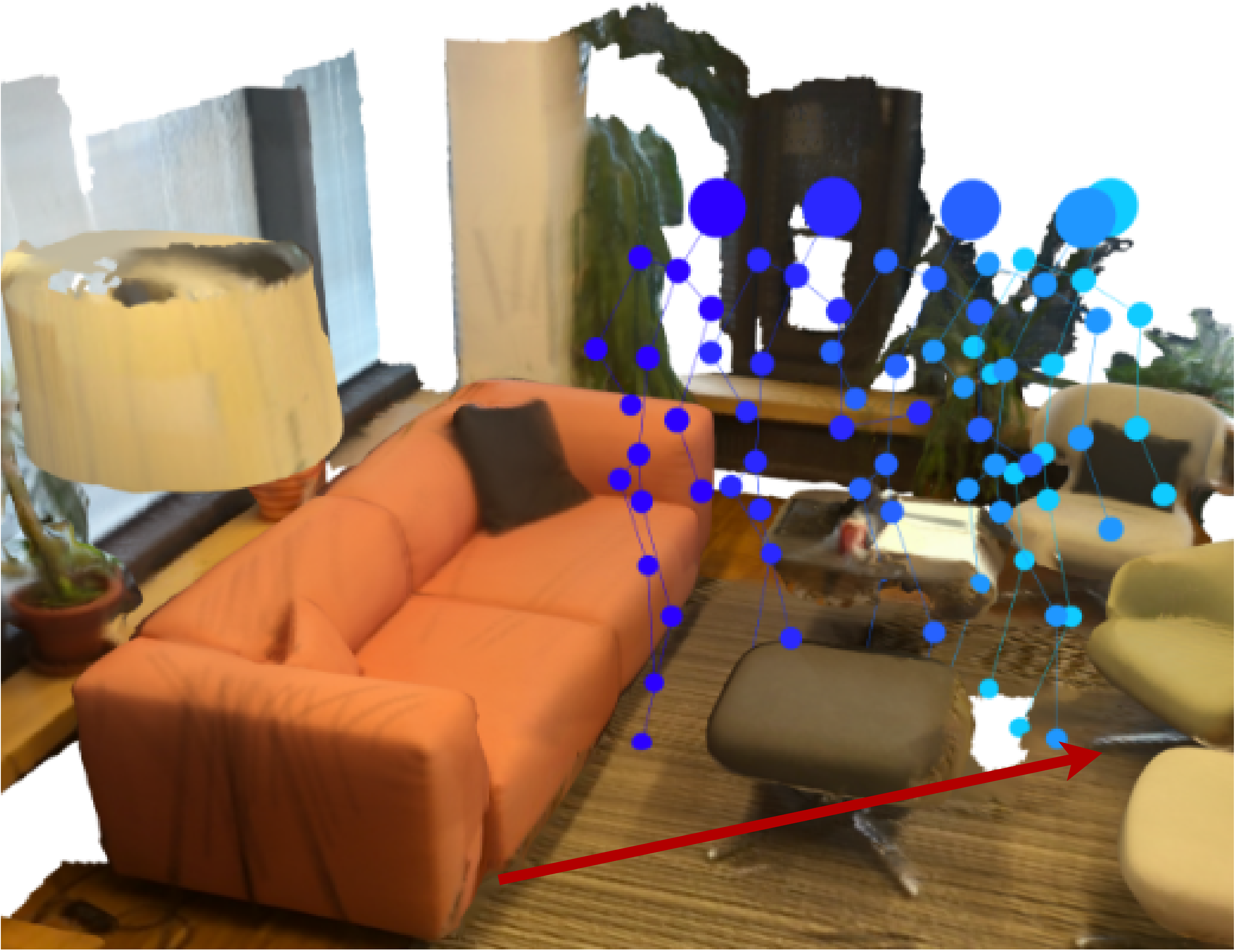}\\
\includegraphics[width=0.42\linewidth]{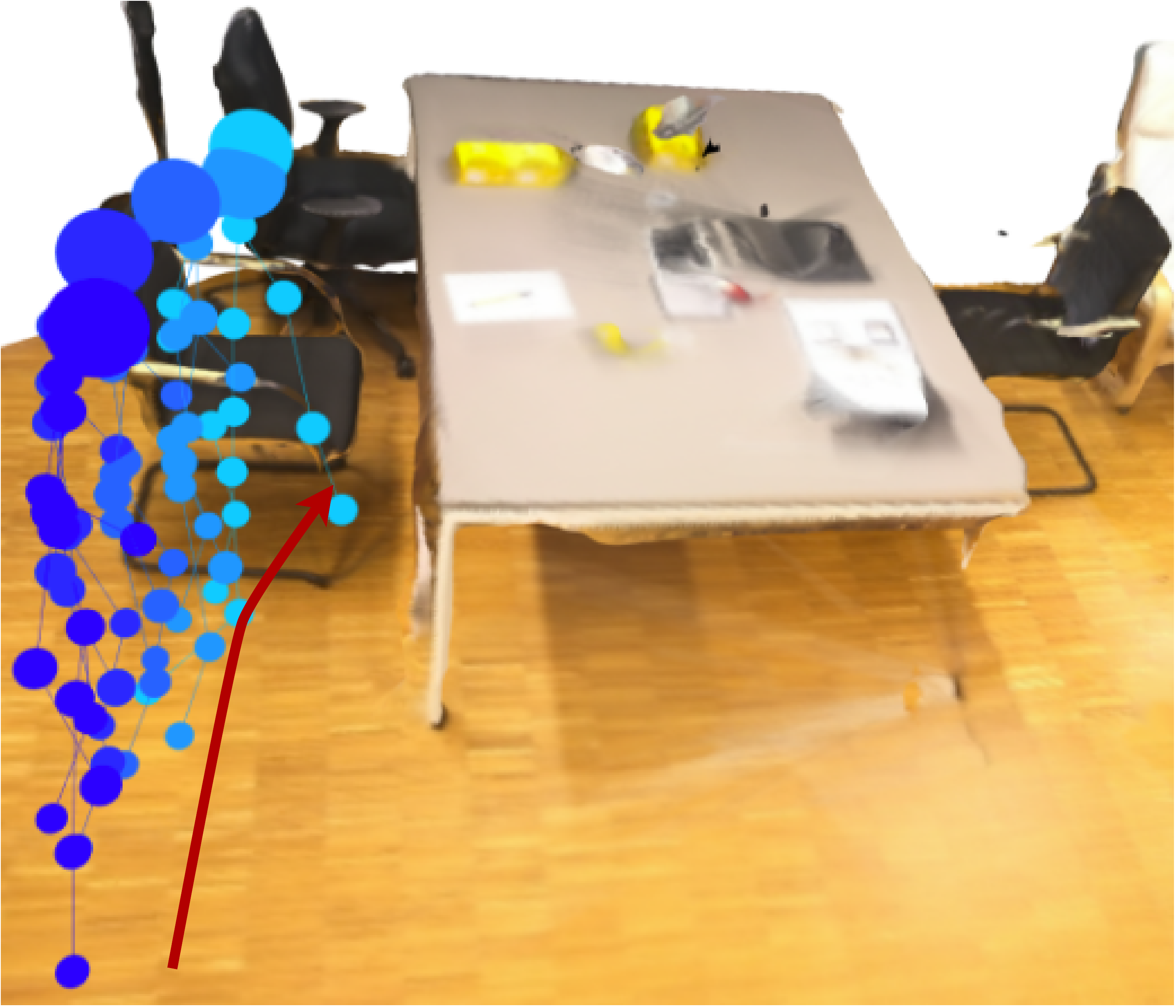} &
\includegraphics[width=0.42\linewidth]{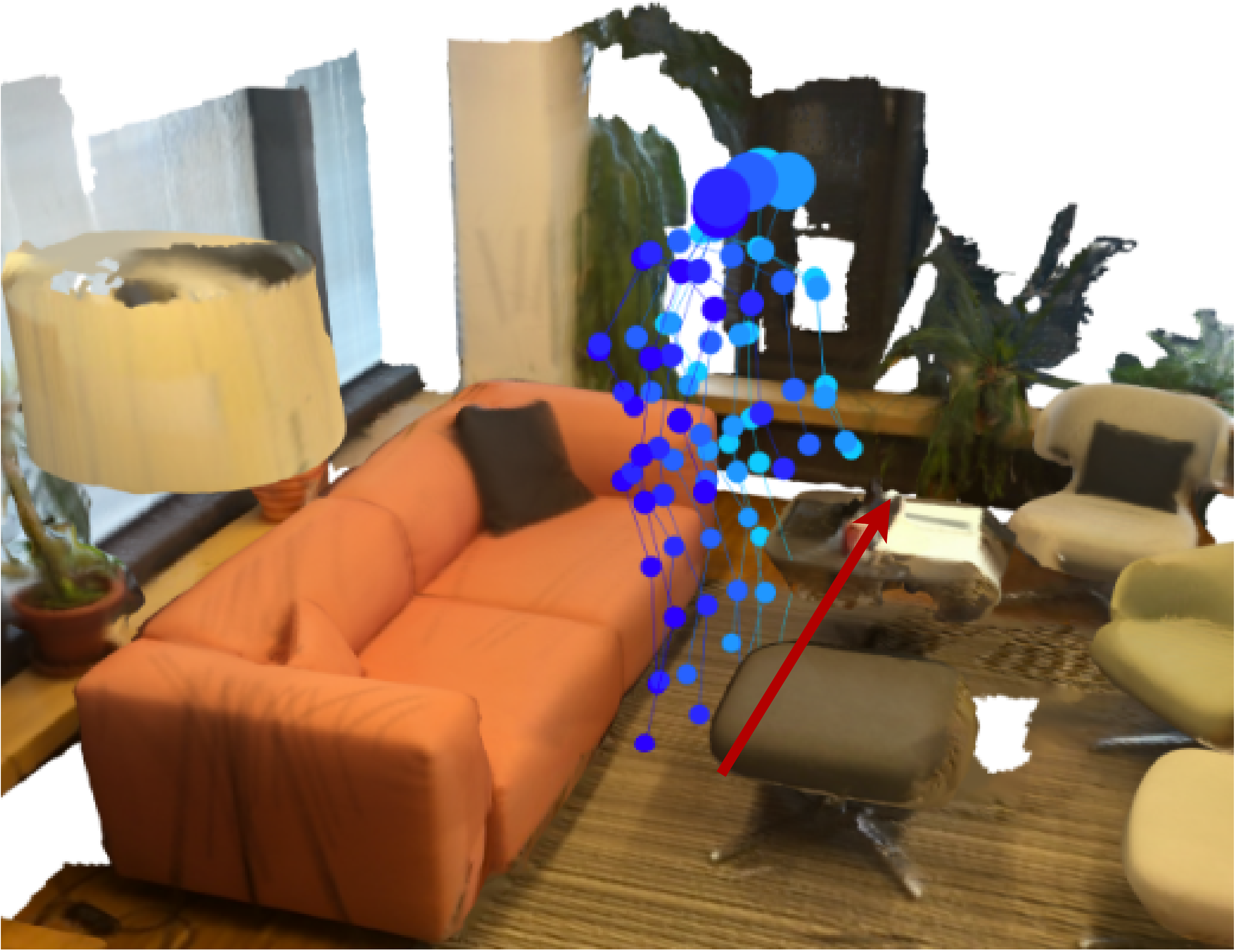} \\
(a) & (b)
\end{tabular}
\caption{\textbf{Visualization on PROX}. Our method can synthesize diverse and compatible human motions in these scenes, although the environment in this dataset is more complex than GTA-IM.}\label{fig:prox_results}
\vspace{-6mm}

\end{center}
\end{figure}

\begin{table}[t]
\centering
\small
\caption{
User studies on both GTA-IM and PROX for different methods. M: separately generative framework of human motion; G: supervision of geometry discriminators; GT: ground-truth human motions. The user study score is written as \emph{mean(std)}.
}
\label{tab:user study}
\setlength{\tabcolsep}{1mm}{
\begin{tabular}{l | c c c c}
\shline
\multirow{2}*{Method}& \multicolumn{2}{c }{GTA-IM} &  \multicolumn{2}{c}{PROX}\\

 & Scene $\uparrow$ & Human $\uparrow$ & Scene $\uparrow$ & Human $\uparrow$\\
 \hline
 CSGN & 2.66(0.74) & 2.92(0.84) & 2.46(0.82) & 2.52(0.73)\\
 CSGN + \emph{M} &  3.26(0.76) & 3.32(0.73)&  3.12(0.88) & 3.23(0.71) \\
 CSGN +  \emph{M} + \emph{G}& \textbf{3.56}(0.72) & \textbf{3.53}(0.68)  & \textbf{3.36}(0.74) & \textbf{3.32}(0.66) \\
 \hline
 \emph{GT} & 4.51 (0.49)& 4.45(0.49) & 4.13(0.62) & 4.14(0.61)\\
\shline
\end{tabular}}
\end{table}

\subsection{Qualitative Results}
At last, we show more qualitative results on the GTA-IM dataset and the PROX dataset in Figure~\ref{fig:gta_results} and Figure~\ref{fig:prox_results}. From these figures, our method can synthesize diverse and compatible motions in both synthetic and real-world environments. Especially, the first row in Figure~\ref{fig:gta_results} indicates that our method can synthesize specific motions on chairs or stairs. And in the second row of this figure, we can also find out that the synthesized human motion turning around to avoid obstacles. The third and fourth rows demonstrate that our method can synthesize the diverse motions with same and different end points from the same start points. More qualitative results are in the following video\footnote{Please refer to \url{https://www.youtube.com/watch?v=XfA3QWcV0ik&t=10s} for demo videos.}.

%% file: Sections/section_conclusion.tex
\section{Conclusion}
In this paper, we propose a novel scene-aware generative framework to model the distribution of human motion in the given scene.
We at first decouple this distribution into the distribution of trajectories in the scene and the distribution of body movements based on the scene and trajectory. And a projection discriminator and a context discriminator are further introduced into our framework to encourage the compatibility between human motions and the scene.
On two large-scale datasets, we demonstrate the effectiveness of the proposed generative framework
which is able to synthesize human motions that are not only diverse in both trajectories and body movements,
but also coherent with both structure and semantics of the given scene.

\paragraph{Acknowledgments} This work is supported by the Collaborative Research Grant from SenseTime(CUHK Agreement No.TS1712093) and the General Research Fund (GRF) of Hong Kong (No.14205719).

%% file: Sections/appendix.tex
\section{Appendix}
\subsection{Network Structure}
We outline our network architectures in this section. Specifically, all branches in the following tables are modified by removing all the normalization and activation layers. We utilized the LeakeyReLu~\cite{he2015delving} function before all convolution layers and the Batch Normalization~\cite{ioffe2015batch} layer after each convolution. For convenience, we show all these networks which are deployed on GTA-IM dataset~\cite{cao2020long}.
    Our codes will be released to ensure reproducibility.

Firstly, the network structure of the trajectory generator and the pose generator are demonstrated in Table~\ref{trajectory generator} and Table~\ref{pose generator}. As shown in these tables, the condition features of these generators are first repeated to the same spatial shape as the noise and then concatenated to this noise as the input.  Moreover, human motion can be synthesized by aligning the sampled trajectories and pose sequences.

Besides, we utilize trajectory discriminator and pose discriminator to keep the synthesized trajectories and pose sequences smooth and continuous. The network structures of these two discriminators are shown in Table~\ref{trajectory discriminator} and Table~\ref{pose discriminator} respectively. As our generators, the condition features are first repeated to the same spatial-temporal shape and then concatenated to the sampled trajectory and pose sequence. Specifically, we directly concatenate the initial pose to the sampled pose sequence as the first frame, and the temporal shape of this sequence is changed from 64 to 65.

At last, the structure of our projection discriminator and context discriminator are also illustrated in Table~\ref{projection discriminator} and Table~\ref{context discriminator}. The input of the projection discriminator is pre-processed like our pose discriminator, and the input of our context discriminator is the sequence of the cropped relative depth maps. For computing efficiency, we keep the cropped relative depth maps by 8 frame intervals, and the sequence length of this geometry context is 9.
\subsection{More Details for Motion Synthesis}
Firstly, besides the given scene, our framework needs the initial pose as the condition for motion synthesis.
To properly evaluate our framework, we utilize ground-truth initial poses during training and test, as input scene images provided by datasets contain the initial poses. Directly sampling different initial poses may lead to inconsistency between the poses and the ground-truth scene images. It is restricted by datasets rather than our framework. 
Besides, all experiments in our paper are for 64 frame motion sequences. It is noticed that our model does not limit the length of synthesized motions. Our model can generate longer sequences using a sliding window. However, in this way, it may be harder to control the overall smoothness of the entire sequence caused by the CSGN~\cite{yan2019convolutional}. Therefore, a more direct way is to train the framework with longer motions. In the supplemented video, we demonstrate the qualitative results of longer synthesized motions from our framework. The more efficient way to synthesize longer videos will be the future work for researchers.
\begin{table}[t]
	\centering
	\caption{The network structure of our \textbf{Trajectory Generator}. We first repeat scene context and initial pose 6 times by the third dimension and concatenate them to the sampled noise. $Conv$ means convolution operator, and $Up$ means the temporal-wise upsampling operator.}\label{trajectory generator}
	\begin{tabular}{llll}
	\shline
	Block & Operation & Input & Output\\
	\hline
	\multirow{3}*{In} & Noise  & (256, 1, 6) & \multirow{3}*{(569, 1, 6)}\\
		& Scene Context & (256, 1, 1)\\ 
		& Initial Pose & (57, 1, 1)&\\

	\hline
	(1) & $Conv$  & (569, 1, 6) & (512, 1, 4) \\
	(2) & $Conv + Up$ & (512, 1, 4) & (256, 1, 8) \\
	(3) & $Conv + Up$ & (256, 1, 8) & (128, 1, 16) \\
	(4) & $Conv + Up$ & (128, 1, 16) & (64, 1, 32) \\
	(5) & $Conv + Up$ & (64, 1, 32) & (32, 1, 64) \\
	\hline
	Out & $Conv + Tanh$ & (32, 1, 64) & (3, 1, 64) \\
	\shline
	\end{tabular}
\end{table}

\begin{table}[t]
	\centering
	\caption{The network structure of our \textbf{Pose Generator}. We first repeat scene context, initial pose, and the sampled trajectory 6 times by the third dimension and concatenate them to the sampled noise. $Conv_{st}$ means graph convolution operator~\cite{yan2019convolutional} and $Up$ means the temporal-wise upsampling operator.}\label{pose generator}
	\begin{tabular}{llll}
	\shline
	Block & Operation & Input & Output\\
	\hline
	\multirow{4}*{In} & Noise  & (1024, 1, 6) & \multirow{4}*{(1529, 1, 6)}\\
		& Scene Context & (256, 1, 1)\\ 
		& Trajectory & (192, 1, 1) \\
		& Initial Pose & (57, 1, 1)&\\

	\hline
	(1) & $Conv_{st} $  & (1529, 1, 6) & (512, 5, 4) \\
	(2) & $Conv_{st} + Up$ & (512, 5, 4) & (256, 5, 8) \\
	(3) & $Conv_{st} + Up$ & (256, 5, 8) & (128, 11, 16) \\
	(4) & $Conv_{st} + Up$ & (128, 11, 16) & (64, 11, 32) \\
	(5) & $Conv_{st} + Up$ & (64, 11, 32) & (32, 19, 64) \\
	\hline
	Out & $Conv_{st} + Tanh$ & (32, 19, 64) & (3, 19, 64) \\
	\shline
	\end{tabular}
\end{table}

\begin{table}[t]
	\centering

	\caption{The network structure of our \textbf{Projection Discriminator}.We first repeat scene context to the same spatial-temporal shape of the 2D human motion and then concatenate the scene context to this sequence as input..$Conv_{st}$ means graph convolution operator~\cite{yan2019convolutional}, $Down$ means the temporal-wise downsampling operator, and $Pool$ means the global average pooling.}\label{projection discriminator}
	\begin{tabular}{llll}
	\shline
	Block & Operation & Input & Output\\
	\hline
	\multirow{3}*{In} & 2D Motion & (2, 19, 65) & \multirow{3}*{(258, 19, 65)}\\
		& Scene Context & (256, 1, 1)\\

	\hline
	(1) & $Conv_{st}$  & (258, 19, 65) &  (64, 11, 65) \\
	(2) & $Conv_{st} + Down$ & (64, 11, 65) & (64, 11,  32) \\
	(3) & $Conv_{st} + Down$ & (64, 11, 32) & (128, 5, 16) \\
	(4) & $Conv_{st} + Down$ & (128, 5, 16) & (256, 5, 8) \\
	(5) & $Conv_{st} + Down$ & (256, 5, 8) & (512, 1, 4) \\
	\hline
	Out & $Conv + Pool$ & (512, 1, 4) & (512, 1, 1) \\
	\shline
	\end{tabular}
\end{table}

\begin{table}[t]
	\centering

	\caption{The network structure of our \textbf{Context Discriminator}. We define the cropped relative depth maps which are guided by trajectory as the input geometry context of this branch. $Conv$ means convolution operator, $Down$ means the temporal-wise downsampling operator, and $Pool$ means the global average pooling.}\label{context discriminator}
	\begin{tabular}{llll}
	\shline
	Block & Operation & Input & Output\\
	\hline
	In & Geometry Context & (9, 72, 128) & (9, 72, 128) \\
	\hline
	(1) & $Conv+Down$  & (9, 72, 128) &  (64, 36, 64) \\
	(2) & $Conv+Down$  & (64, 36, 64) &  (128, 18, 32) \\
	(3) & $Conv+Down$  & (128, 18, 32) &  (256, 9, 16) \\
	\hline
	Out & $Conv + Pool$ & (256, 9, 16) & (512, 1, 1) \\
	\shline
	\end{tabular}
\end{table} 

\begin{table}[t]
	\centering

	\caption{The network structure of our \textbf{Trajectory Discriminator}. We first repeat scene context and initial pose 64 times by the third dimension and concatenate them to the sampled trajectory. $Conv$ means convolution operator, $Down$ means the temporal-wise downsampling operator, and $Pool$ means the global average pooling.}\label{trajectory discriminator}
	\begin{tabular}{llll}
	\shline
	Block & Operation & Input & Output\\
	\hline
	\multirow{3}*{In} & Trajectory & (3, 1, 64) & \multirow{3}*{(316, 1, 64)}\\
		& Scene Context & (256, 1, 1)\\ 
		& Initial Pose & (57, 1, 1)\\

	\hline
	(1) & $Conv$  & (316, 1, 64) &  (64, 1, 64) \\
	(2) & $Conv + Down$ & (64, 1, 64) & (64, 1,  32) \\
	(3) & $Conv + Down$ & (64, 1, 32) & (128, 1, 16) \\
	(4) & $Conv + Down$ & (128, 1, 16) & (256, 1, 8) \\
	(5) & $Conv + Down$ & (256, 1, 8) & (512, 1, 4) \\
	\hline
	Out & $Conv + Pool$ & (512, 1, 4) & (512, 1, 1) \\
	\shline
	\end{tabular}
\end{table}

\begin{table}[t]
	\centering

	\caption{The network structure of our \textbf{Pose Discriminator}. We first repeat scene context to the same spatial-temporal shape of the pose sequence and then concatenate all condition features to this sequence as input. $Conv_{st}$ means graph convolution operator~\cite{yan2019convolutional}, $Down$ means the temporal-wise downsampling operator, and $Pool$ means the global average pooling.}\label{pose discriminator}
	\begin{tabular}{llll}
	\shline
	Block & Operation & Input & Output\\
	\hline
	\multirow{3}*{In} & Trajectory & (3, 1, 65) & \multirow{3}*{(262, 19, 65)}\\
		& Scene Context & (256, 1, 1)\\ 
		& Pose& (3, 19, 65)\\

	\hline
	(1) & $Conv_{st}$  & (262, 19, 65) &  (64, 11, 65) \\
	(2) & $Conv_{st} + Down$ & (64, 11, 65) & (64, 11,  32) \\
	(3) & $Conv_{st} + Down$ & (64, 11, 32) & (128, 5, 16) \\
	(4) & $Conv_{st} + Down$ & (128, 5, 16) & (256, 5, 8) \\
	(5) & $Conv_{st} + Down$ & (256, 5, 8) & (512, 1, 4) \\
	\hline
	Out & $Conv + Pool$ & (512, 1, 4) & (512, 1, 1) \\
	\shline
	\end{tabular}
\end{table}